\documentclass{article}

\usepackage[main, final]{neurips_2026}

\usepackage[utf8]{inputenc} 
\usepackage[T1]{fontenc}    
\usepackage{hyperref}       
\usepackage{url}            
\usepackage{booktabs}       
\usepackage{amsfonts}       
\usepackage{nicefrac}       
\usepackage{microtype}      
\usepackage{xcolor}         
\usepackage{amsmath}
\usepackage{multirow}
\usepackage{wrapfig}
\usepackage{graphicx}  
\usepackage{amsmath}   
\usepackage{graphicx}
\usepackage{subcaption}
\usepackage{xcolor}
\newcommand{\pos}[1]{\textcolor{green!50!black}{#1}}

\title{VC-OPD: Visual Counterfactual On-Policy Distillation for Grounded Vision-Language Reasoning}

\author{%
  Jiahe Feng \\
  College of Intelligence and Computing \\
  Tianjin University \\
  Tianjin, China \\
  \texttt{fjh3475227047@tju.edu.cn} \\
}

\begin{document}

\maketitle

\begin{abstract}
GRPO-style reinforcement learning has advanced vision-language reasoning, but its sequence-level rewards provide limited credit assignment for individual reasoning tokens. On-policy distillation (OPD) addresses this issue by providing dense supervision on student-generated trajectories, yet standard OPD assigns similar importance to all teacher-preferred tokens, including template phrases and visually irrelevant continuations. We introduce Visual Counterfactual On-Policy Distillation (VC-OPD), a grounding-aware distillation method that prioritizes tokens whose correctness depends on image-specific evidence. After a teacher-supervised warm start that stabilizes response formats and reduces noisy rollouts, VC-OPD performs on-policy distillation with two teacher evaluations for each student trajectory: one under the original image and one under a counterfactual image with corrupted instance-specific evidence. The original-image teacher distribution determines the distillation target, while the original--counterfactual teacher gap estimates token-level visual dependence and softly reweights the loss. To avoid unstable response-level update scales, VC-OPD further applies mass-preserving normalization to the token weights. This design preserves the dense supervision of OPD while shifting learning capacity toward visually grounded reasoning tokens. Experiments on multimodal reasoning benchmarks show that VC-OPD improves visual reasoning performance and produces more effective token-level distillation.
\end{abstract}

\section{Introduction}

Large vision-language models (VLMs) have achieved strong progress in multimodal reasoning, yet post-training them remains substantially harder than post-training text-only language models. Recent reinforcement and preference-based methods, including PPO, GRPO, multimodal RLHF, reinforcement fine-tuning, and verifiable-reward optimization, have improved mathematical and long-chain reasoning~\citep{schulman2017ppo,shao2024deepseekmath,sun2023llavarlhf,yu2023rlhfv,liu2025visualrft,shen2025vlmr1,zhang2025r1vl,tan2025reasonrft}. However, VLM reasoning has an additional prerequisite: the model must first identify and use the relevant visual evidence. This has motivated perception-aware post-training methods, such as corrupted-image objectives, perception-first curricula, self-rewarding decomposition, and grounded visual exploration~\citep{wang2025papo,chen2025pebr,li2025visionsr1,sarch2025vigorl}. These studies indicate that effective VLM post-training should not only optimize final answers, but also assign learning signals to tokens that are genuinely grounded in the image.

On-policy distillation (OPD) is a natural framework for such fine-grained supervision. Instead of relying only on sequence- or step-level rewards, OPD samples trajectories from the student and asks a teacher to score the same trajectories, providing dense token-level feedback under the student's own state distribution~\citep{agarwal2024gkd}. Recent theoretical analyses further interpret OPD as KL-regularized token-level policy optimization, where the teacher--student likelihood ratio acts as a dense reward-like signal~\citep{yang2026gopd,ko2026reopold}. Nevertheless, token-level density does not imply grounding awareness. Standard OPD treats all teacher-preferred tokens as equally distillable, although many tokens in VLM responses are generic templates, discourse markers, or language-prior continuations rather than image-dependent content.

\begin{figure*}[t]
    \centering
    \includegraphics[width=0.95\textwidth]{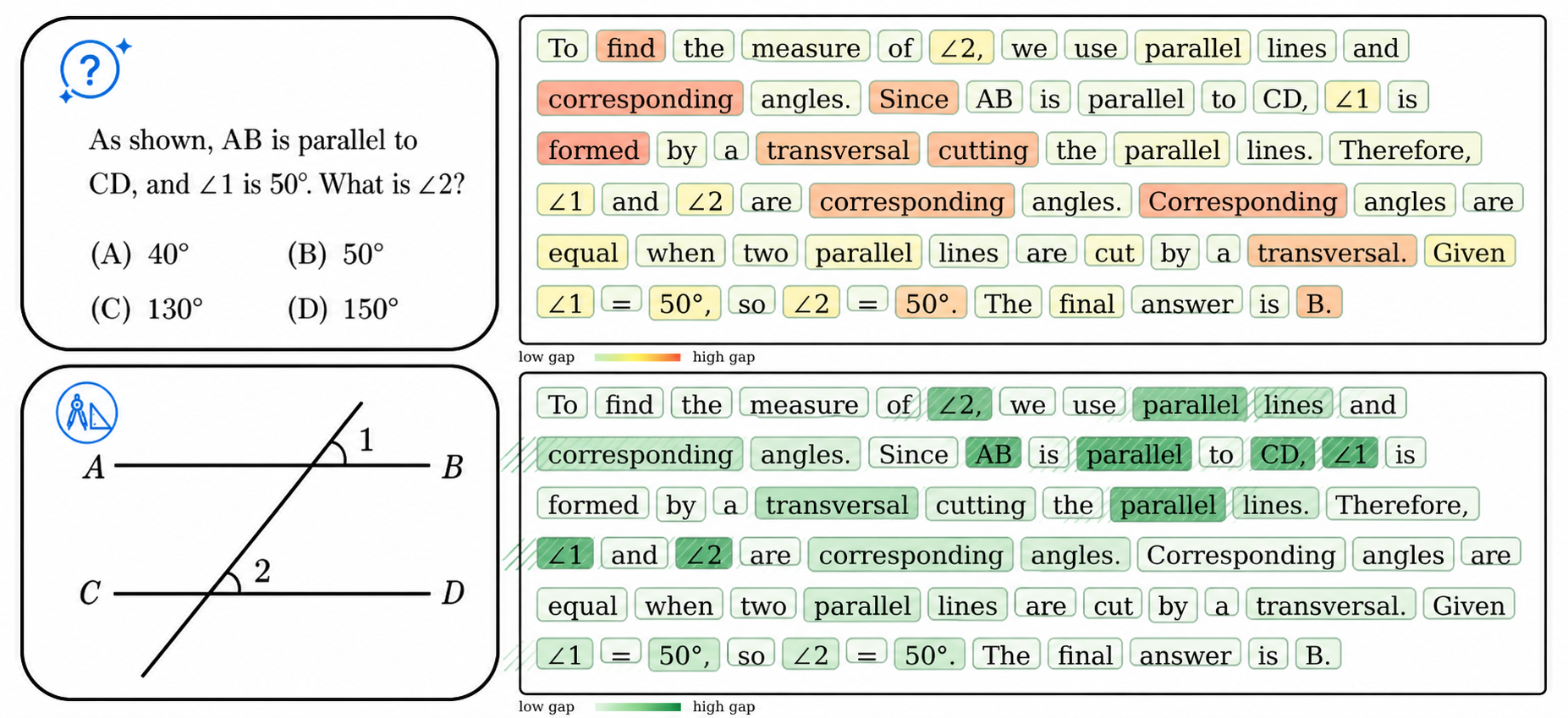}
    \caption{
    Token-level grounding visualization.
    Standard student--teacher gaps highlight both visual and generic tokens, whereas factual--counterfactual teacher gaps more selectively emphasize image-dependent tokens.
    }
    \label{fig:token_grounding_visualization}
\end{figure*}

This limitation is illustrated in Figure~\ref{fig:token_grounding_visualization}. The student--teacher gap used by standard OPD highlights both visually meaningful tokens and generic reasoning expressions, revealing where the student differs from the teacher but not whether the discrepancy is grounded in the image. In contrast, a factual--counterfactual teacher gap, obtained by comparing teacher scores under the original image and a visually weakened counterfactual image, more selectively emphasizes tokens tied to visual evidence, such as angle symbols, parallel-line relations, and answer-relevant quantities.

\begin{figure*}[t]
    \centering
    \begin{subfigure}[t]{0.32\textwidth}
        \centering
        \includegraphics[width=\linewidth]{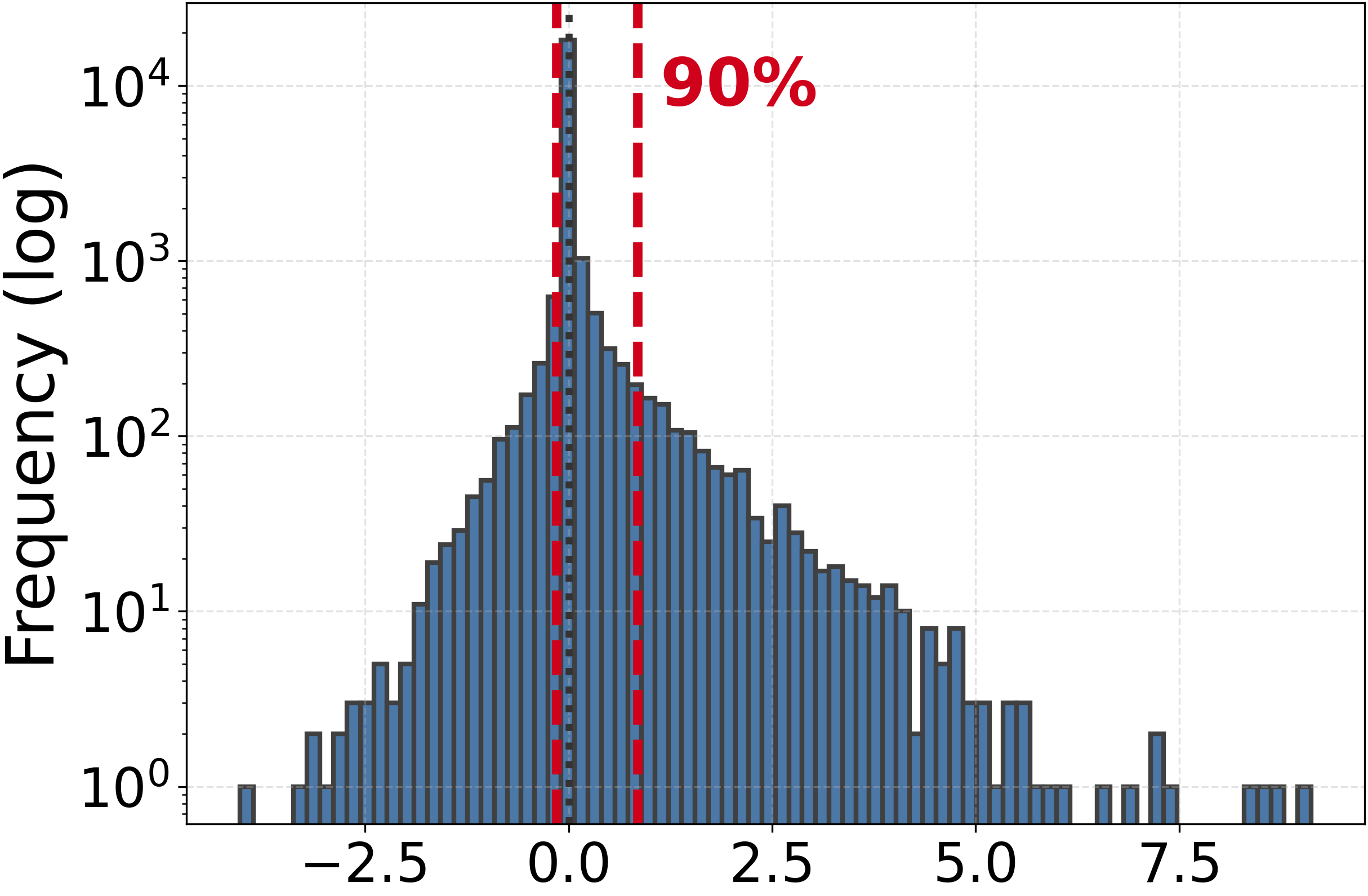}
        \caption{Student--teacher gap}
        \label{fig:student_teacher_gap}
    \end{subfigure}
    \hfill
    \begin{subfigure}[t]{0.32\textwidth}
        \centering
        \includegraphics[width=\linewidth]{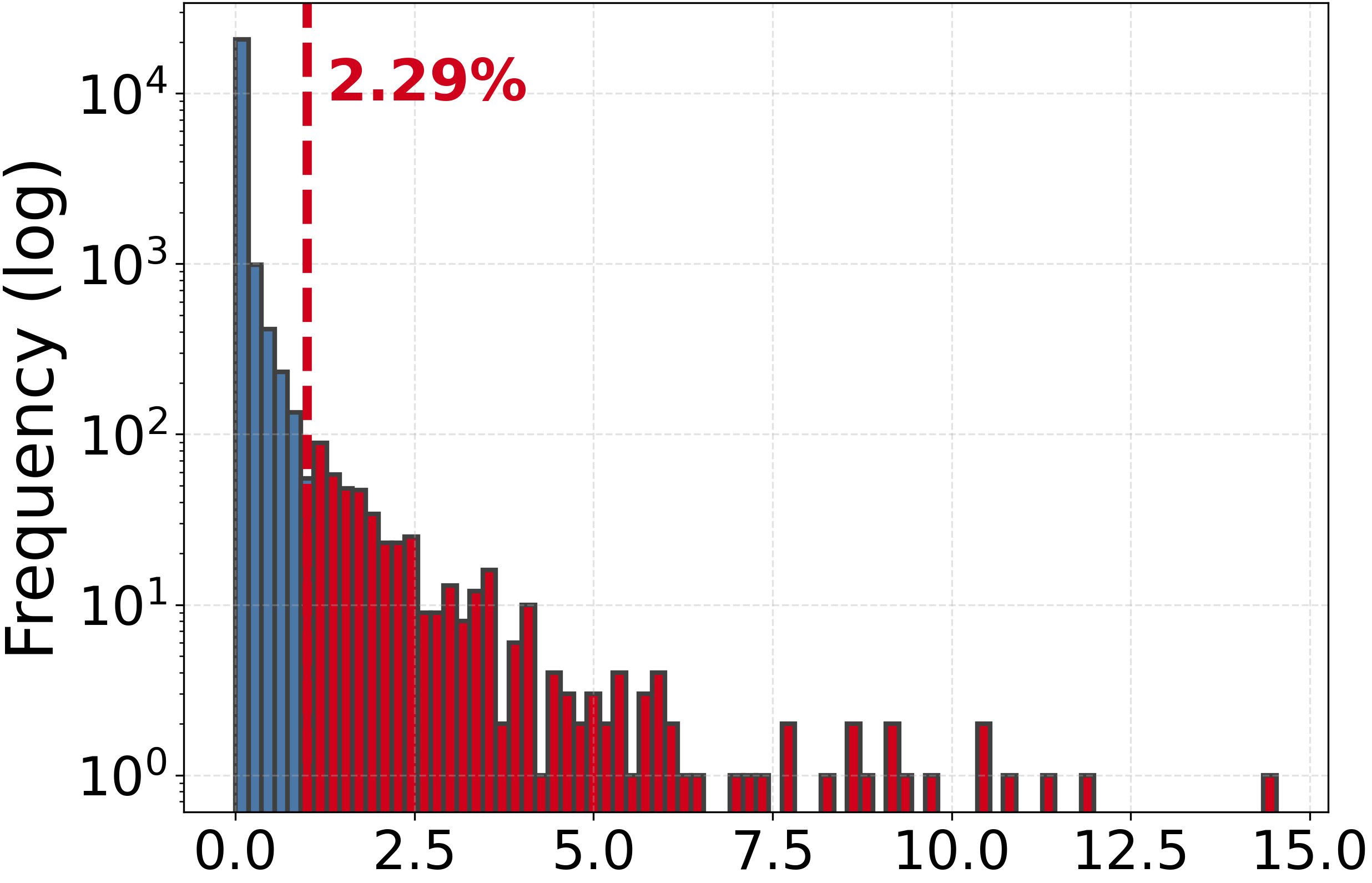}
        \caption{Factual--counterfactual gap}
        \label{fig:factual_counterfactual_gap}
    \end{subfigure}
    \hfill
    \begin{subfigure}[t]{0.32\textwidth}
        \centering
        \includegraphics[width=\linewidth]{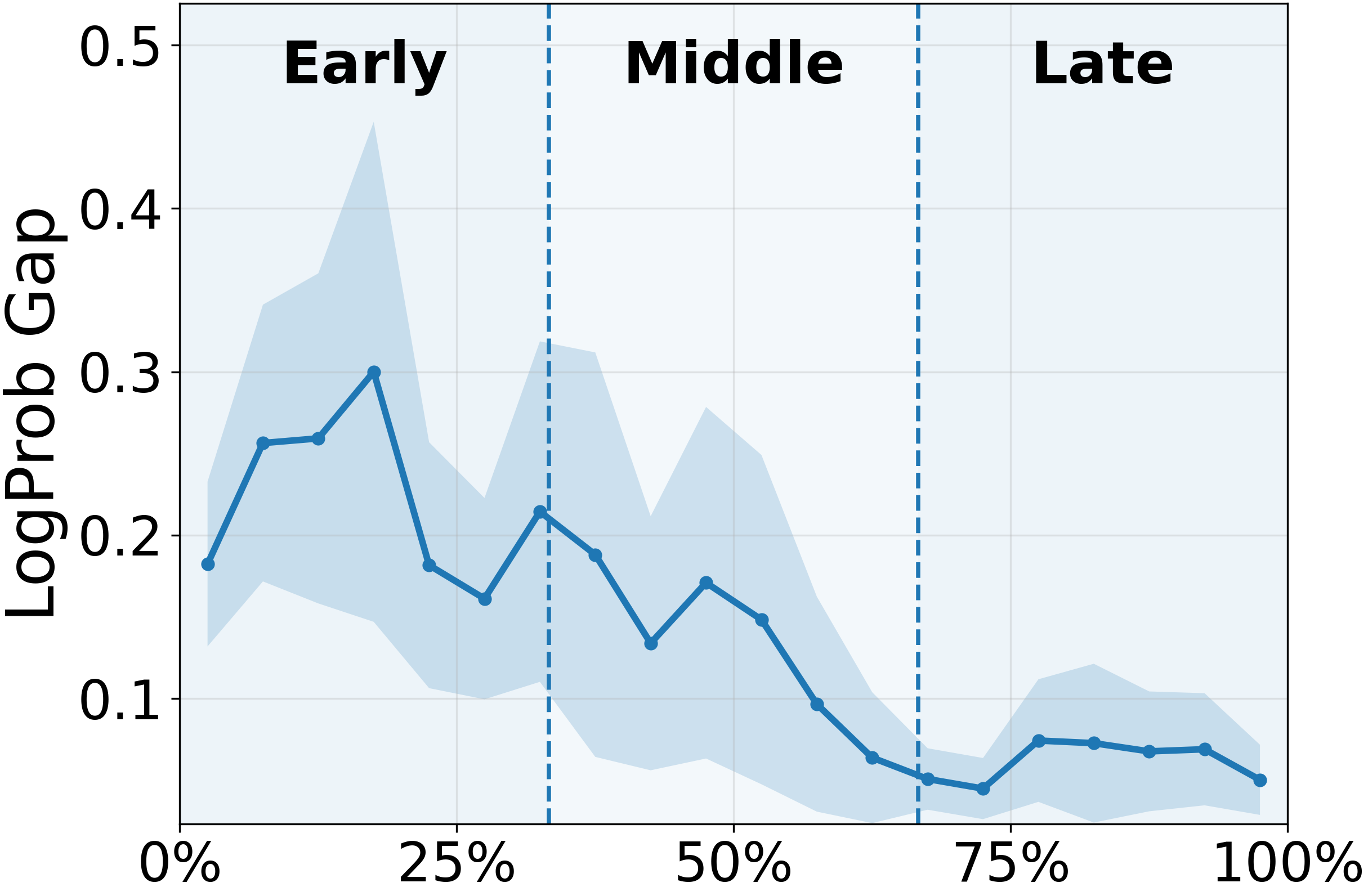}
        \caption{Position-wise visual gap}
        \label{fig:positionwise_visual_gap}
    \end{subfigure}
    \caption{
    Token-level signal statistics.
    Standard OPD gaps are often small, while factual--counterfactual gaps reveal sparse and position-dependent visual grounding signals.
    }
    \label{fig:token_signal_analysis}
\end{figure*}

Token-level statistics provide the same message at scale. As shown in Figure~\ref{fig:token_signal_analysis}, student--teacher gaps are heavily concentrated near zero, suggesting that dense OPD supervision often yields weak effective gradients. By contrast, large factual--counterfactual gaps appear on only a sparse subset of tokens, consistent with the structure of VLM reasoning: only some tokens encode objects, quantities, spatial relations, or diagram-specific information, while many others mainly serve as linguistic scaffolding. The position-wise pattern further shows that visual dependence is stronger in early perception-related positions and weaker in later reasoning or answer-formatting stages. These observations motivate a grounding-aware OPD objective that preserves dense supervision while reallocating update strength toward visually dependent tokens.

We propose \emph{Visual Counterfactual On-Policy Distillation} (VC-OPD), a grounding-aware OPD framework for vision-language reasoning. VC-OPD decouples the \emph{direction} and \emph{magnitude} of token-level updates. For each student-generated trajectory, the factual teacher under the original image provides the teacher--student discrepancy that determines the update direction. The same teacher is then evaluated under a counterfactual image where instance-specific visual evidence is weakened. The resulting factual--counterfactual log-probability gap estimates each token's dependence on the original visual input and is converted into a soft grounding weight that modulates the update magnitude. In this way, VC-OPD retains the dense token-level supervision of OPD while concentrating learning on tokens that are more likely to encode visual perception and grounding.

VC-OPD uses a simple two-stage pipeline. First, we warm-start the student with teacher-generated responses to stabilize response format and reduce noisy on-policy rollouts. Second, we perform visual counterfactual on-policy distillation: the student rolls out under the real image, the teacher scores the same trajectory under both factual and counterfactual visual conditions, and the resulting visual-dependence score reweights the token-level OPD objective. We further apply mass-preserving normalization to preserve the relative emphasis on visually grounded tokens while keeping the overall token weight stable across responses.

Our contributions are threefold:
\begin{itemize}
    \item We identify a grounding-related limitation of standard OPD for VLM reasoning: the teacher--student discrepancy captures model disagreement but does not explicitly distinguish image-dependent tokens from generic linguistic continuations.
    \item We introduce the factual--counterfactual teacher log-probability gap as a token-level proxy for visual dependence, enabling grounding-aware reweighting of student-generated trajectories.
     \item We formulate VC-OPD with a direction--magnitude separation: the factual teacher--student discrepancy determines the update direction, while the counterfactual signal modulates its magnitude through soft weighting and mass-preserving normalization.
\end{itemize}


\section{Related Work}

\subsection{Reinforcement learning for vision-language models}

Reinforcement learning has become an important post-training paradigm for vision-language models (VLMs), building on the success of policy optimization in language reasoning. Modern VLM RL inherits trust-region and proximal policy methods such as TRPO and PPO, as well as critic-free variants such as GRPO, which has shown strong performance in mathematical reasoning~\citep{schulman2015trpo,schulman2017ppo,shao2024deepseekmath}. These optimization paradigms have been extended to multimodal alignment and reasoning through RLHF, reward modeling, preference alignment, and reinforcement fine-tuning~\citep{sun2023llavarlhf,yu2023rlhfv,zhang2025mmrlhf,ji2025saferlhfv,liu2025visualrft,shen2025vlmr1,zhang2025r1vl,tan2025reasonrft,liu2025visualarft}.

A growing line of work further argues that visual perception is a central bottleneck in VLM RL. PAPO introduces perception-aware policy optimization by comparing model behavior under original and corrupted visual inputs; PeBR proposes a two-stage curriculum that improves perception before reasoning; Vision-SR1 decomposes visual reasoning into perception and language reasoning with self-rewarding signals; ViGoRL grounds intermediate reasoning steps to visual coordinates~\citep{wang2025papo,chen2025pebr,li2025visionsr1,sarch2025vigorl}. These methods show that successful multimodal reasoning requires models to use instance-specific visual evidence rather than relying on language priors or generic reasoning templates. 
However, most existing VLM RL methods still optimize sequence-level rewards, step-level rewards, or auxiliary perception losses, and do not provide dense teacher-guided token-level credit assignment. Our method addresses this gap by introducing visual counterfactual signals into on-policy distillation.


\subsection{On-policy distillation for reasoning models}

Knowledge distillation transfers capabilities from large teachers to smaller students, but classical distillation and supervised fine-tuning usually optimize the student on fixed teacher-provided trajectories, which can introduce distribution mismatch~\citep{hinton2015distilling,kim2016seqkd}. On-policy distillation (OPD) reduces this mismatch by sampling trajectories from the student and querying the teacher on those trajectories, thereby aligning supervision with the student's own rollout distribution~\citep{agarwal2024gkd,yang2026gopd,ko2026reopold}.

Recent work has provided a stronger theoretical and practical foundation for OPD. GKD formalizes on-policy distillation for autoregressive models~\citep{agarwal2024gkd}. G-OPD interprets OPD as dense KL-constrained policy optimization and generalizes the reference-policy design~\citep{yang2026gopd}. REOPOLD further shows that the teacher--student likelihood ratio can be viewed as a token-level reward and improves training efficiency through relaxed token selection and reward clipping~\citep{ko2026reopold}. Other OPD variants study context distillation, reasoning-prefix supervision, and entropy-aware objectives~\citep{ye2026opcd,zhang2026prefixopd,jin2026entropy}. Compared with sparse reward RL, OPD provides dense token-level supervision, which is attractive for long-chain reasoning.

While effective for text reasoning, existing OPD methods typically treat all generated tokens as distillable. This assumption is suboptimal for VLM reasoning, where many tokens are structural markers, template phrases, or language-prior continuations, and only a subset is truly supported by visual evidence. We extend the generalized OPD perspective to multimodal reasoning by using a counterfactual teacher to estimate token-level visual dependence and reweighting OPD updates toward visually grounded tokens.

\section{Method}

\subsection{Preliminaries: on-policy distillation as token-level RL}

We consider a vision-language reasoning setting with textual input $x$, visual input $v$, and autoregressive output sequence $y=(y_1,\dots,y_T)$. The student policy is denoted by $\pi_\theta(y_t\mid x,v,y_{<t})$, and the rollout policy $\mu$ is typically the current student or a lagged copy. During training, we sample trajectories on-policy under the real visual input:
\begin{equation}
y \sim \mu(\cdot \mid x,v).
\end{equation}

On-policy distillation (OPD) differs from conventional offline distillation in one essential aspect: the teacher does not supervise teacher-forced trajectories, but instead evaluates trajectories that are actually generated by the student itself. As a result, the distillation signal is aligned with the student's own state distribution, which reduces the train--test mismatch that commonly arises in offline knowledge distillation. This property is particularly important for reasoning tasks, where small deviations in early tokens may lead the student to entirely different downstream trajectories.

Recent work has shown that OPD can be interpreted through the lens of token-level policy optimization. Let $s_t=(x,v,y_{<t})$ denote the autoregressive state and $a_t=y_t$ the sampled token. If we measure the discrepancy between the student and the teacher with the reverse KL divergence,
\begin{equation}
D_{\mathrm{RKL}}\!\left(\pi_\theta(\cdot\mid s_t)\,\|\,\pi_{\mathrm{tea}}(\cdot\mid s_t)\right)
=
\mathbb{E}_{a_t\sim \pi_\theta}
\left[
\log \pi_\theta(a_t\mid s_t)-\log \pi_{\mathrm{tea}}(a_t\mid s_t)
\right],
\end{equation}
then maximizing the negative reverse KL induces the token-level learning signal
\begin{equation}
r_t
=
\log \pi_{\mathrm{tea}}(a_t\mid s_t)-\log \pi_\theta(a_t\mid s_t).
\label{eq:token_reward}
\end{equation}
This quantity can be viewed as a dense per-token reward: if the teacher assigns higher probability to the sampled token than the student, the reward is positive, encouraging the student to increase its probability; otherwise the reward becomes small or negative. In this sense, OPD can be understood as a form of dense RL where the reward is not provided by an external environment, but is instead induced by the discrepancy between the teacher and the student on student-generated trajectories.

Under this view, a broad class of OPD-style methods can be written as weighted log-likelihood optimization:
\begin{equation}
\mathcal{L}(\theta)
=
-
\mathbb{E}_{y\sim\mu}
\left[
\sum_{t=1}^{T}
\mathrm{sg}(A_t)\log \pi_\theta(y_t\mid x,v,y_{<t})
\right],
\label{eq:weighted_distill}
\end{equation}
where $A_t$ is an advantage-like token weight and $\mathrm{sg}(\cdot)$ denotes stop-gradient. Therefore, once OPD is viewed as token-level RL, the central modeling question becomes how to define $A_t$.

For text-only reasoning, this weight mainly reflects how strongly the teacher prefers a continuation. In multimodal reasoning, however, teacher preference alone is insufficient: a token may be preferred either because it is genuinely grounded in visual evidence or because it is a generic reasoning phrase, a structural template token, or a continuation supported only by language priors. Our goal is therefore to redesign the token-level weight so that student updates are concentrated on tokens that are truly supported by visual evidence.

\begin{figure*}[t]
    \centering
    \includegraphics[width=\textwidth]{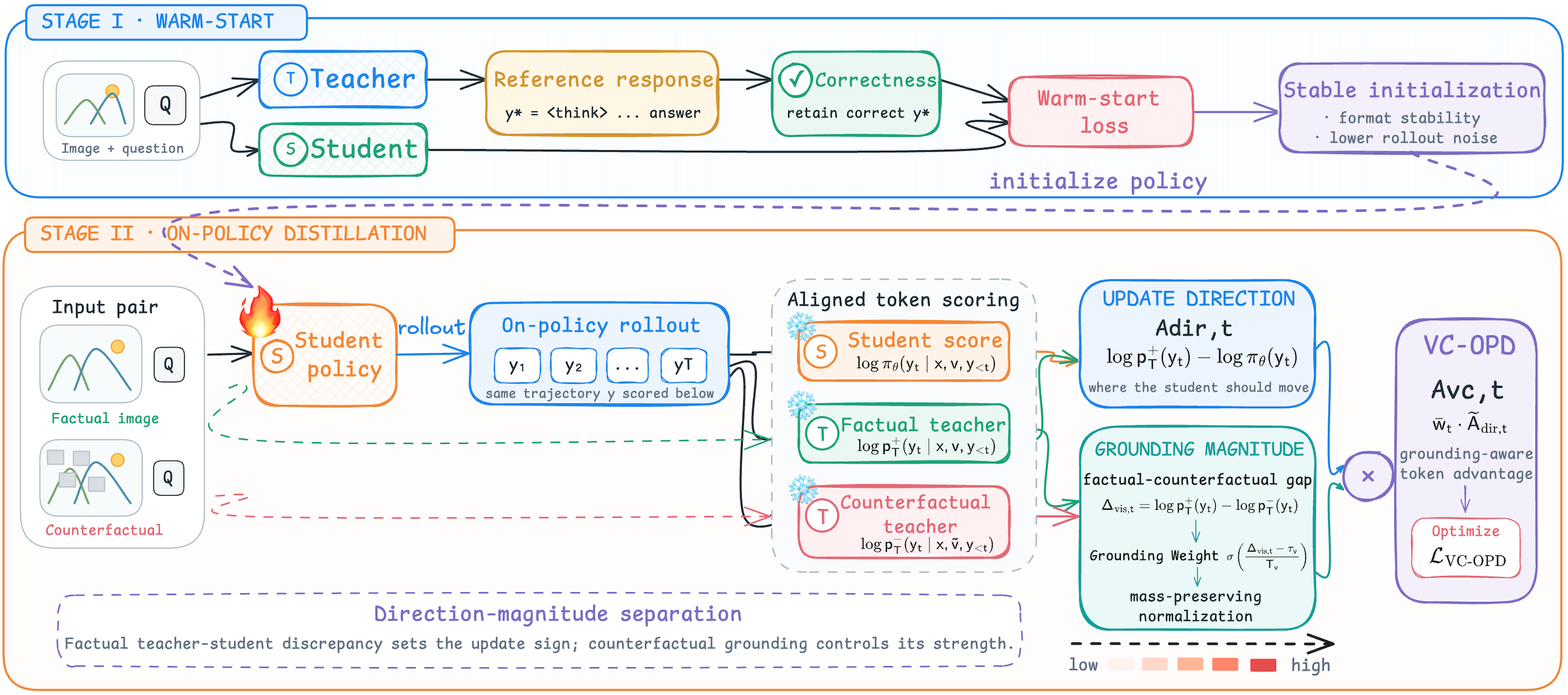}
    \caption{
    Overview of VC-OPD.
    The framework first warm-starts the student with filtered teacher-generated reasoning responses.
    During visual counterfactual on-policy distillation, student rollouts are scored by the teacher under factual and counterfactual visual inputs.
    The factual teacher provides the update direction, while the factual--counterfactual gap produces grounding-aware token weights that modulate the final token-level advantage.
    }
    \label{fig:vcopd_overview}
\end{figure*}

\subsection{Two-stage training}

Our training consists of two stages. The first stage provides a stable initialization, and the second stage performs visual counterfactual on-policy distillation.

\paragraph{Stage I: warm-start.}
Before entering on-policy training, we first warm-start the student using teacher-generated reference responses $y^\star$. Given $(x,v)$, the teacher produces a structured target response, and the student is optimized with standard autoregressive supervision:
\begin{equation}
\mathcal{L}_{\mathrm{warm}}(\theta)
=
-
\mathbb{E}_{(x,v,y^\star)\sim\mathcal{D}}
\left[
\sum_{t=1}^{T^\star}
\log \pi_\theta(y_t^\star\mid x,v,y^\star_{<t})
\right].
\label{eq:warm_start}
\end{equation}
The purpose of this stage is not to solve the final problem by imitation alone, but to place the student in a reasonable response regime before on-policy optimization. In practice, this reduces rollout noise, stabilizes response format, and makes the subsequent teacher feedback more informative.

\paragraph{Stage II: visual counterfactual on-policy distillation.}
After warm-start, the student generates its own response trajectory under the real image,
\begin{equation}
y \sim \mu(\cdot\mid x,v).
\end{equation}
We then evaluate the same trajectory with the teacher under two visual conditions: the original image and a counterfactual image in which instance-specific evidence has been weakened. The factual branch determines \emph{where} the student should move, while the counterfactual branch determines \emph{how strongly} each token should be updated.

\subsection{Counterfactual teacher scoring and grounding-aware weighting}

For each sampled token $y_t$, we define the factual teacher score and the counterfactual teacher score as
\begin{equation}
p_T^{+}(y_t)=p_T(y_t\mid x,v,y_{<t}), \qquad
p_T^{-}(y_t)=p_T(y_t\mid x,\tilde v,y_{<t}),
\label{eq:teacher_scores}
\end{equation}
where $\tilde v=c(v)$ is a corrupted visual input produced by a visual perturbation operator $c(\cdot)$. The role of $c(\cdot)$ is not to destroy the entire input, but to weaken instance-level visual evidence while preserving the general input format, so that the teacher remains in a comparable prediction regime.

Based on these two teacher evaluations, we compute two quantities. The first is a direction term,
\begin{equation}
A_t^{\mathrm{dir}}
=
\log p_T^{+}(y_t)-\log \mu(y_t\mid x,v,y_{<t}),
\label{eq:direction}
\end{equation}
which compares the factual teacher and the student on the sampled token. This term determines the basic update direction: when the student underestimates a token relative to the factual teacher, the token receives a positive update signal; when the student already overestimates it, the update is weakened or reversed.

The second is a visual dependence score,
\begin{equation}
\Delta_t^{\mathrm{vis}}
=
\log p_T^{+}(y_t)-\log p_T^{-}(y_t),
\label{eq:visual_gap}
\end{equation}
This metric quantifies the shift in teacher token support under visual perturbation. We use this difference as a proxy for token-level visual dependence, not a rigorous causal estimate of image effects. A positive value means the original factual image yields stronger teacher support for the token, while near-zero values show insensitivity to visual perturbations. The score hinges on whether the perturbation removes task-relevant visual evidence while preserving sufficient context.


Rather than hard-selecting tokens with a binary rule, we convert the visual dependence score into a continuous soft weight:
\begin{equation}
w_t
=
\sigma\!\left(
\frac{\Delta_t^{\mathrm{vis}}-\tau_v}{T_v}
\right),
\label{eq:soft_weight}
\end{equation}
where $\sigma(\cdot)$ is the sigmoid function, $\tau_v$ is a visual threshold, and $T_v$ is a temperature parameter. This design preserves continuity during training: highly grounded tokens receive larger weights, while weakly grounded tokens are downweighted but not necessarily discarded.

A practical issue is that the number and strength of visually grounded tokens may vary substantially across samples, which can lead to unstable update scales if $w_t$ is used directly. To stabilize optimization, we adopt a mass-preserving normalization:
\begin{equation}
\bar w_t
=
w_t \cdot
\frac{N}{\sum_{j=1}^{T} w_j+\epsilon},
\qquad
N=\sum_{t=1}^{T} m_t,
\label{eq:normalized_weight}
\end{equation}
where $m_t$ is the response mask and $N$ is the number of valid response tokens. This normalization preserves the relative emphasis across tokens while keeping the total effective weight approximately constant for each response, thereby preventing large variations in update magnitude across batches.

\subsection{Final VC-OPD objective}

We combine the direction term and the normalized grounding weight to define the final token-level advantage:
\begin{equation}
A_t^{\mathrm{VC}}
=
\bar w_t \cdot \tilde A_t^{\mathrm{dir}},
\label{eq:final_advantage}
\end{equation}
where $\tilde A_t^{\mathrm{dir}}$ denotes a normalized version of $A_t^{\mathrm{dir}}$ for numerical stability. The key design principle is that the factual teacher--student discrepancy determines the direction of the update, while the counterfactual branch modulates only its magnitude according to visual grounding strength.

The final stage-II objective is
\begin{equation}
\mathcal{L}_{\mathrm{VC\text{-}OPD}}(\theta)
=
-
\mathbb{E}_{y\sim\mu(\cdot\mid x,v)}
\left[
\sum_{t=1}^{T}
\mathrm{sg}\!\left(A_t^{\mathrm{VC}}\right)
\log \pi_\theta(y_t\mid x,v,y_{<t})
\right].
\label{eq:final_objective}
\end{equation}
This objective retains the on-policy nature of OPD while making the token-level update explicitly grounding-aware. Tokens that are both under-modeled by the student and strongly supported by real visual evidence receive larger updates; tokens that are mainly driven by language priors are correspondingly suppressed.

\section{Experiments}
\subsection{Experimental setup}

We use Qwen2.5-VL~\citep{bai2025qwen25vl} as the primary backbone for our main
experiments. Unless otherwise specified, Qwen2.5-VL-3B-Instruct is the student,
while larger Qwen2.5-VL models serve as teachers. All methods start from the
same student checkpoint and are trained on ViRL39K~\citep{wang2025vlrethinker},
which contains approximately 39K verifiable visual-language reasoning examples.
VC-OPD uses a two-stage procedure consisting of teacher-supervised warm-start
and visual counterfactual on-policy distillation. We further evaluate Qwen3-VL and InternVL3.5 to test cross-backbone generalization, with additional details provided in Appendix \ref{app:Qwen3-VL} and \ref{app:InternVL3.5}.

We evaluate seven multimodal reasoning benchmarks covering mathematical,
geometry, multi-disciplinary, logical, and counting tasks: Geometry3K~\citep{lu2021intergps},
MathVista~\citep{lu2023mathvista}, MathVerse~\citep{zhang2024mathverse},
We-Math~\citep{qiao2024wemath}, MMMU-Pro~\citep{yue2024mmmupro},
LogicVista~\citep{xiao2024logicvista}, and Super-CLEVR Counting~\citep{li2023superclevr}.
All results use exact match after answer extraction, with the average reported
as the unweighted mean across benchmarks.

Unless otherwise specified, we use the @8 setting with inference temperature
$1.0$. All experiments are conducted on four NVIDIA A800 GPUs. Detailed
hyperparameters and implementation details are provided in Appendix~\ref{app:training_details}.

\begin{table*}[t]
\centering
\small
\setlength{\tabcolsep}{4.2pt}
\renewcommand{\arraystretch}{1.12}
\caption{
Main results on multimodal reasoning benchmarks. We report exact-match accuracy under the @8 setting with inference temperature $1.0$. 
Avg. denotes the unweighted average over all listed benchmarks.
}
\label{tab:main_results}
\resizebox{\textwidth}{!}{
\begin{tabular}{lcccccccc}
\toprule
\textbf{Model}
& \textbf{Geo3K}
& \textbf{MVer.}
& \textbf{MVist.}
& \textbf{WeM.}
& \textbf{MMMU}
& \textbf{Logic}
& \textbf{Count}
& \textbf{Avg.} \\
\midrule

Qwen2.5-VL-3B-Instruct
& 24.96 & 38.08 & 58.50 & 53.22 & 24.50 & 34.72 & 51.71 & 40.81 \\
\midrule

\multicolumn{9}{l}{\textit{7B teacher $\rightarrow$ 3B student}} \\
\quad + GRPO
& 30.52 & 40.60 & 59.04 & 59.10 & 25.86 & 38.34 & 57.81 & 44.47 \\
\quad + PAPO
& 30.95 & 41.50 & 62.38 & 60.09 & 27.11 & 38.67 & 62.56 & 46.18 \\
\quad + RKL
& 32.12 & 40.75 & 62.08 & 60.23 & 28.40 & 39.41 & 63.50 & 46.64 \\
\quad + VC-OPD
& \textbf{34.35} & \textbf{41.90} & \textbf{64.10} & \textbf{61.46}
& \textbf{32.20} & \textbf{42.40} & \textbf{65.68} & \textbf{48.87} \\
\quad $\Delta_{\mathrm{RKL}}$ (\%)
& \pos{+6.94} & \pos{+2.82} & \pos{+3.25} & \pos{+2.04}
& \pos{+13.38} & \pos{+7.59} & \pos{+3.43} & \pos{+4.78} \\
\midrule

\multicolumn{9}{l}{\textit{32B teacher $\rightarrow$ 3B student}} \\
\quad + RKL
& 45.09 & 43.25 & 65.50 & 64.56 & 31.24 & 41.52 & 65.80 & 50.99 \\
\quad + VC-OPD
& \textbf{57.58} & \textbf{44.40} & \textbf{65.72} & \textbf{65.74}
& \textbf{33.50} & \textbf{42.90} & \textbf{67.24} & \textbf{53.87} \\
\quad $\Delta_{\mathrm{RKL}}$ (\%)
& \pos{+27.70} & \pos{+2.66} & \pos{+0.34} & \pos{+1.83}
& \pos{+7.23} & \pos{+3.32} & \pos{+2.19} & \pos{+5.64} \\
\midrule
Qwen2.5-VL-7B-Instruct
& 36.80 & 43.90 & 66.40 & 64.80 & 33.80 & 43.50 & 64.90 & 50.59 \\
\midrule

\multicolumn{9}{l}{\textit{32B teacher $\rightarrow$ 7B student}} \\
\quad + RKL
& 48.70 & 47.40 & 69.70 & 66.30 & 36.20 & 46.00 & 69.00 & 54.76 \\
\quad + VC-OPD
& \textbf{55.40} & \textbf{48.90} & \textbf{70.90} & \textbf{68.10}
& \textbf{38.80} & \textbf{48.60} & \textbf{71.20} & \textbf{57.41} \\
\quad $\Delta_{\mathrm{RKL}}$ (\%)
& \pos{+13.76} & \pos{+3.17} & \pos{+1.72} & \pos{+2.71}
& \pos{+7.18} & \pos{+5.65} & \pos{+3.19} & \pos{+4.85} \\
\bottomrule
\end{tabular}}
\vspace{2pt}
\end{table*}

\subsection{Main results}

Table~\ref{tab:main_results} reports the main results on multimodal reasoning benchmarks.
We adopt Qwen2.5-VL-3B-Instruct and Qwen2.5-VL-7B-Instruct as the student models,
with Qwen2.5-VL-7B and Qwen2.5-VL-32B serving as teacher models.
For the Qwen2.5-VL-3B student, we evaluate both 7B- and 32B-teacher settings,
while the Qwen2.5-VL-7B student is distilled from the 32B teacher.
In the 7B-teacher setting with the 3B student, we compare VC-OPD with GRPO, PAPO,
and reverse-KL on-policy distillation (RKL).
For all distillation settings, RKL serves as the direct on-policy distillation baseline,
and $\Delta_{\mathrm{RKL}}$ denotes the relative improvement of VC-OPD over RKL.

VC-OPD consistently improves over RKL across all evaluated student--teacher settings.
With the 7B teacher and the Qwen2.5-VL-3B student, VC-OPD improves the average accuracy
from $46.64$ to $48.87$, yielding a $2.23$-point absolute gain and a $4.78\%$
relative improvement.
With the 32B teacher and the same 3B student, VC-OPD further improves the average
accuracy from $50.99$ to $53.87$, corresponding to a $2.87$-point absolute gain and a
$5.64\%$ relative improvement.
For the Qwen2.5-VL-7B student, the base model achieves an average accuracy of $50.59$,
while distillation from the 32B teacher improves the RKL performance from $54.76$ to
$57.41$, corresponding to a $4.85\%$ relative improvement.
The consistent gains across different student and teacher scales demonstrate that
VC-OPD remains effective as the student model capacity increases, while preserving
the advantages of grounding-aware token-level distillation.

\subsection{Ablation study}

We conduct ablation studies under the 7B-teacher setting to isolate the effect of each component in VC-OPD. 
Starting from vanilla on-policy distillation, we progressively examine the contribution of warm-start training, counterfactual teacher scoring, soft visual reweighting, and mass-preserving normalization. 

\begin{table*}[t]
\centering
\small
\setlength{\tabcolsep}{6pt}
\renewcommand{\arraystretch}{1.15}
\caption{
Ablation study under the 7B-teacher setting. 
Warm, CF, Soft, and Mass denote warm-start, counterfactual teacher scoring, soft visual reweighting, and mass-preserving normalization, respectively.
}
\label{tab:ablation}
\begin{tabular}{lccccccc}
\toprule
\textbf{Variant}
& \textbf{Warm}
& \textbf{CF}
& \textbf{Soft}
& \textbf{Mass}
& \textbf{Math Avg.}
& \textbf{Vision Avg.}
& \textbf{Overall Avg.} \\
\midrule
Vanilla OPD
& -- & -- & -- & -- 
& 48.80 & 43.77 & 46.64 \\
w/o warm-start
& -- & \checkmark & \checkmark & \checkmark 
& 49.37 & 45.55 & 47.73 \\
w/o counterfactual teacher
& \checkmark & -- & -- & \checkmark 
& 49.42 & 44.63 & 47.37 \\
w/ hard mask
& \checkmark & \checkmark & -- & \checkmark 
& 49.86 & 45.92 & 48.17 \\
w/o mass norm.
& \checkmark & \checkmark & \checkmark & -- 
& 50.05 & 46.10 & 48.36 \\
\midrule
VC-OPD
& \checkmark & \checkmark & \checkmark & \checkmark
& \textbf{50.45} & \textbf{46.76} & \textbf{48.87} \\
\bottomrule
\end{tabular}
\end{table*}

Table~\ref{tab:ablation} shows that each component contributes to the final performance of VC-OPD. 
Vanilla OPD already provides dense token-level supervision, but its performance remains limited because all generated tokens are treated uniformly. 
Adding counterfactual visual scoring improves both math-oriented and visually grounded benchmarks, indicating that token-level visual dependence is a useful signal for multimodal distillation.

Warm-start training also matters: removing it lowers the overall average from $48.87$ to $47.73$, suggesting that stable initial rollouts are important before applying on-policy teacher feedback. 
Replacing soft reweighting with a hard mask further degrades performance, showing that continuous token weighting is preferable to binary token selection. 
Finally, removing mass-preserving normalization reduces the overall average to $48.36$, confirming that stabilizing the effective token weight is beneficial for robust optimization. 
Overall, the full VC-OPD configuration achieves the best performance, supporting the design choice of combining counterfactual visual dependence with stable soft token-level reweighting.

\subsection{Training dynamics}

\begin{figure*}[t]
    \centering
    \begin{subfigure}[t]{0.48\textwidth}
        \centering
        \includegraphics[width=\linewidth]{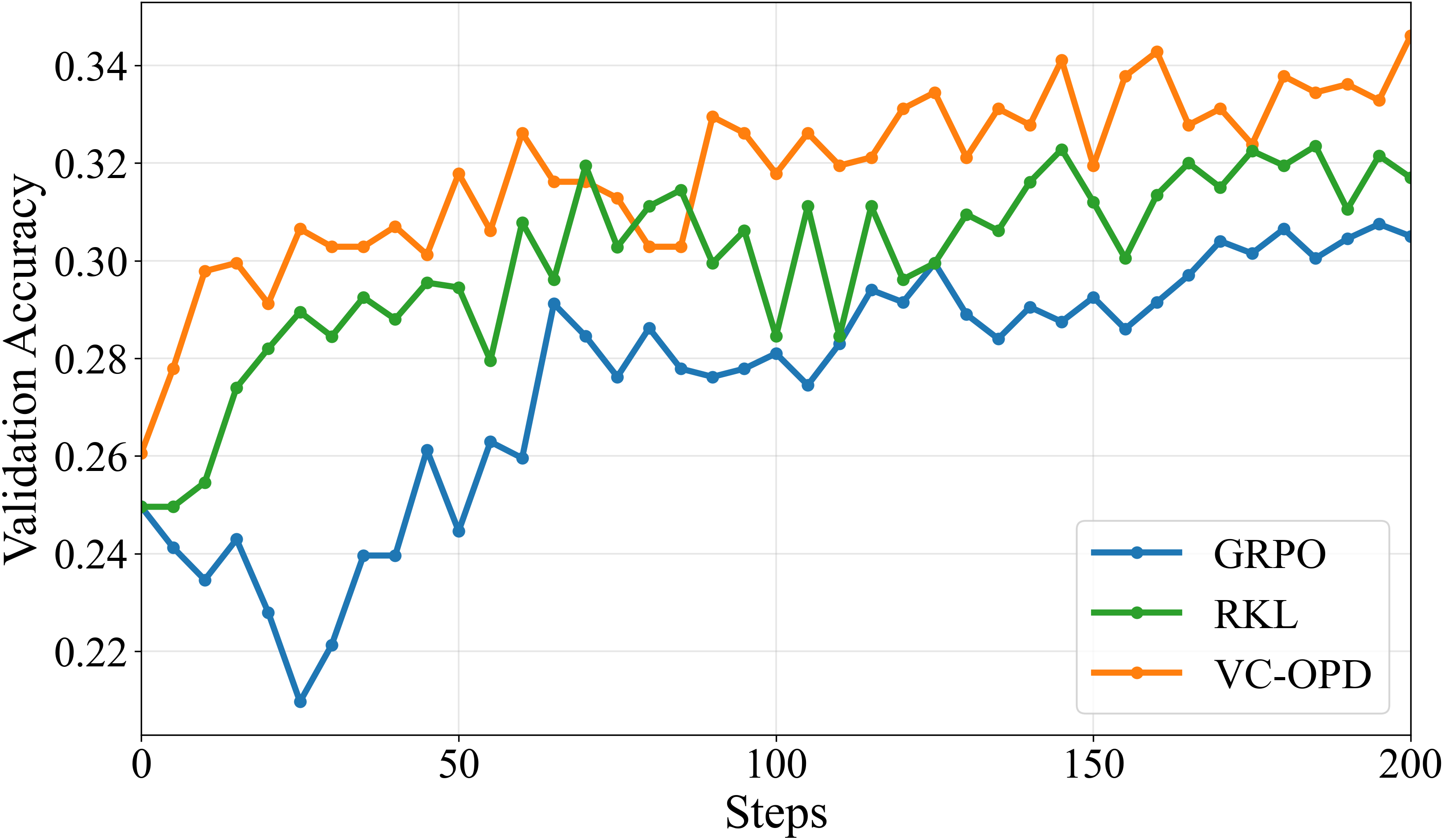}
        \caption{Geo3K validation accuracy}
        \label{fig:geo3k_val_accuracy}
    \end{subfigure}
    \hfill
    \begin{subfigure}[t]{0.48\textwidth}
        \centering
        \includegraphics[width=\linewidth]{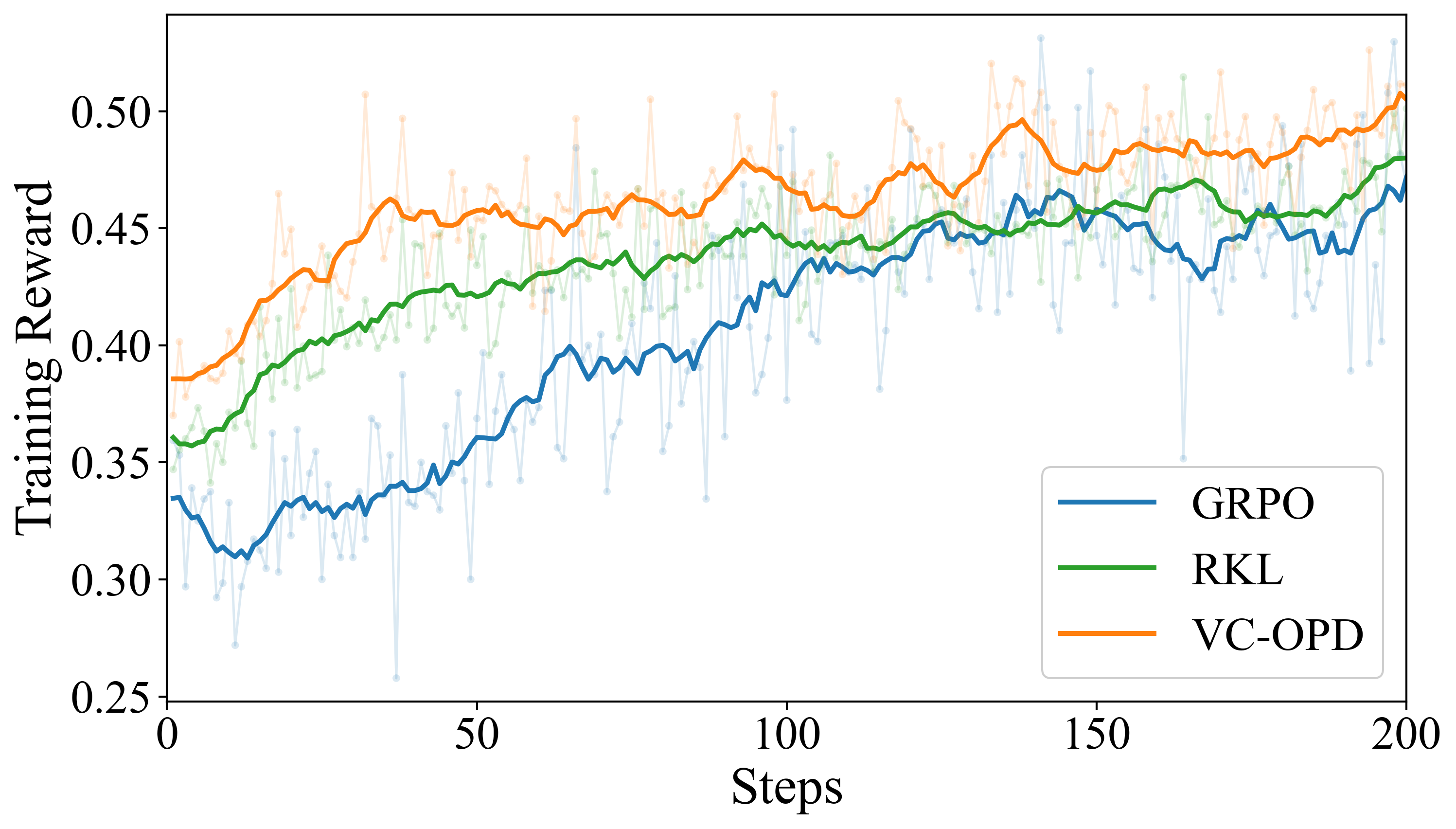}
        \caption{Training reward}
        \label{fig:training_reward}
    \end{subfigure}
    \caption{
    Training dynamics over 200 optimization steps.
    VC-OPD improves faster than GRPO, is more stable than RKL, and achieves the highest validation accuracy and training reward.
    }
    \label{fig:training_dynamics}
\end{figure*}

We further compare the optimization behavior of GRPO, RKL, and VC-OPD under the same 200-step training budget. 
As shown in Table~\ref{tab:training_dynamics} and Figure~\ref{fig:training_dynamics}, both RKL and VC-OPD improve faster than GRPO, suggesting that dense teacher-guided token supervision is more sample-efficient than outcome-level reward optimization. 
VC-OPD further achieves a more stable trajectory than RKL and reaches the best final performance on both Geo3K and LogicVista, indicating that visual counterfactual reweighting improves the effectiveness and stability of token-level distillation.

\begin{table}[t]
\centering
\small
\renewcommand{\arraystretch}{1.15}
\caption{
Step-wise validation accuracy during training. 
We report exact-match accuracy (\%) on Geo3K and LogicVista every 50 optimization steps.
}
\label{tab:training_dynamics}
\begin{tabular*}{0.92\textwidth}{@{\extracolsep{\fill}}lcccccccc@{}}
\toprule
\textbf{Method}
& \multicolumn{4}{c}{\textbf{Geo3K}}
& \multicolumn{4}{c}{\textbf{LogicVista}} \\
\cmidrule(lr){2-5} \cmidrule(lr){6-9}
& \textbf{50} 
& \textbf{100} 
& \textbf{150} 
& \textbf{200}
& \textbf{50} 
& \textbf{100} 
& \textbf{150} 
& \textbf{200} \\
\midrule
GRPO
& 26.10 & 27.75 & 28.90 & 30.52
& 34.20 & 36.30 & 37.45 & 38.34 \\
RKL
& 29.20 & 30.35 & 31.25 & 32.12
& 37.05 & 38.20 & 38.85 & 39.41 \\
VC-OPD
& \textbf{31.80} & \textbf{32.45} & \textbf{33.20} & \textbf{34.35}
& \textbf{39.10} & \textbf{40.55} & \textbf{41.45} & \textbf{42.40} \\
\bottomrule
\end{tabular*}
\end{table}

Compared with GRPO, RKL improves more rapidly because it provides dense token-level teacher feedback instead of relying only on sequence-level rewards. 
However, RKL still applies supervision uniformly across tokens, which may amplify visually irrelevant or template-like continuations. 
VC-OPD addresses this issue by reweighting token updates according to visual counterfactual dependence. 
As a result, it maintains a smoother reward curve and consistently outperforms RKL at later training stages.

\subsection{Hyperparameter analysis}

We analyze the proportion of visually dependent tokens emphasized by the soft visual
weighting function.
The goal is to verify that VC-OPD benefits from a balanced grounding signal rather
than overly dense or overly sparse token reweighting.

\begin{wraptable}{r}{0.49\textwidth}
\vspace{-8pt}
\centering
\small
\caption{
Sensitivity analysis of the visual token ratio.
Default setting is marked with $^\star$.
}
\label{tab:tau_sensitivity}
\setlength{\tabcolsep}{4.0pt}
\renewcommand{\arraystretch}{1.08}
\begin{tabular}{lccc}
\toprule
\textbf{Visual ratio} & \textbf{Math} & \textbf{Vision} & \textbf{Overall} \\
\midrule
20\% & 49.92 & 46.18 & 48.32 \\
10\%$^\star$ & \textbf{50.45} & \textbf{46.76} & \textbf{48.87} \\
5\% & 49.58 & 45.90 & 48.00 \\
\bottomrule
\end{tabular}
\vspace{-6pt}
\end{wraptable}

Table~\ref{tab:tau_sensitivity} shows that a moderate visual-token ratio performs best.
Since visually grounded tokens form only a sparse subset of the response, using a large
ratio such as $20\%$ assigns substantial weight to weakly visual or template-like tokens,
reducing the selectivity of VC-OPD.
In contrast, using only $5\%$ of tokens makes the supervision too sparse, resulting in
insufficient effective updates.
The $10\%$ setting provides a favorable trade-off between grounding selectivity and
optimization density and is therefore used as the default configuration.

\subsection{Alternative Counterfactual Constructions}

We further investigate whether VC-OPD depends on a particular counterfactual
construction.
The counterfactual image is intended to weaken instance-specific visual evidence while
preserving a meaningful visual context.
We compare the default patch dropout with random dropout, Gaussian blur, contiguous
region masking, object-level erasing, saliency-guided corruption, and mismatched-image
counterfactuals under identical training and evaluation settings.

\begin{table*}[t]
\centering
\small
\setlength{\tabcolsep}{4.2pt}
\renewcommand{\arraystretch}{1.10}
\caption{
Comparison of alternative counterfactual visual constructions under the
Qwen2.5-VL-3B student with the Qwen2.5-VL-7B teacher.
The RKL result is shown as the reference baseline.
}
\label{tab:cf_construction}
\vspace{5pt}

\begin{tabular}{llccc}
\toprule
\textbf{Counterfactual}
& \textbf{Configuration}
& \textbf{Math Avg.}
& \textbf{Vision Avg.}
& \textbf{Overall Avg.} \\

\midrule
No counterfactual & RKL & 48.80 & 43.77 & 46.64 \\

\midrule
Random dropout
& --
& 49.86
& 46.20
& 48.29\;
\textcolor{green!50!black}{(+3.54\%)} \\

Patch dropout
& 20\%
& 49.72
& 45.98
& 48.12\;
\textcolor{green!50!black}{(+3.17\%)} \\

Patch dropout
& 60\%
& \textbf{50.45}
& \textbf{46.76}
& \textbf{48.87}\;
\textcolor{green!50!black}{\textbf{(+4.78\%)}} \\

Patch dropout
& 100\%
& 49.20
& 45.55
& 47.64\;
\textcolor{green!50!black}{(+2.14\%)} \\

Gaussian blur
& Kernel $31\times31$, $\sigma=10$
& 50.06
& 46.35
& 48.47\;
\textcolor{green!50!black}{(+3.92\%)} \\

Contiguous region masking
& 60\% masked area
& \underline{50.30}
& \underline{46.55}
& \underline{48.69}\;
\textcolor{green!50!black}{\underline{(+4.40\%)}} \\

Object-level erasing
& Randomly erase object instances
& 49.62
& 45.70
& 47.94\;
\textcolor{green!50!black}{(+2.79\%)} \\

Saliency-guided corruption
& Mask high-saliency regions
& \textbf{50.40}
& \textbf{46.83}
& \textbf{48.87}\;
\textcolor{green!50!black}{\textbf{(+4.78\%)}} \\

Mismatched image
& Random batch permutation
& 47.22
& 41.05
& 44.36\;
\textcolor{red!70!black}{(-4.89\%)} \\

\bottomrule
\end{tabular}

\vspace{2pt}
\end{table*}
Table~\ref{tab:cf_construction} shows that VC-OPD remains effective across a range of
counterfactual constructions.
All alternatives except mismatched-image counterfactuals outperform RKL, indicating
that the proposed objective is not tied to a specific patch-dropout implementation.
Patch dropout at $60\%$ provides a favorable balance between weakening
instance-specific evidence and preserving a meaningful visual prediction context,
and is therefore retained as the default configuration.

The mismatched-image condition performs substantially worse, suggesting that replacing
the original visual context with unrelated evidence does not provide an appropriate
counterfactual for estimating token-level visual dependence.
More targeted constructions, such as object-level erasing and saliency-guided
corruption, are promising but may require additional models or computations, whereas
patch dropout provides a simple and lightweight alternative.

\section{Conclusion}
We introduced VC-OPD, a visual counterfactual extension of on-policy distillation for grounded vision-language reasoning. By comparing teacher scores under factual and counterfactual visual inputs, VC-OPD estimates token-level visual dependence and uses it to reweight dense distillation updates. The factual teacher--student discrepancy determines the update direction, while the counterfactual signal adjusts the update magnitude through soft weighting and mass-preserving normalization. Experiments across multiple student scales and the Qwen2.5-VL, Qwen3-VL, and InternVL3.5 families show consistent improvements over reverse-KL on-policy distillation in the evaluated settings. These findings support grounding-aware token-level credit assignment as a useful complement to general multimodal reasoning objectives.


\section{Limitations}

VC-OPD is primarily designed to improve visually grounded token-level credit
assignment rather than generic text-only reasoning. Its gains may therefore be
more modest on tasks dominated by linguistic logic, symbolic manipulation, or
visually invariant deduction. Moreover, visual dependence is not equivalent to
reasoning importance: some visually invariant transition or reasoning-control
tokens may still play an important role in organizing the reasoning trajectory
and can therefore be underweighted. The factual--counterfactual teacher gap is
also based on visual perturbations rather than a perfect causal intervention,
and the additional counterfactual teacher evaluation introduces computational
overhead. Future work could combine visual-dependence weighting with
complementary signals for reasoning importance and develop more precise and
efficient counterfactual interventions.

\bibliographystyle{unsrtnat}   
\small                         
\bibliography{vc_opd_refs}     

@article{schulman2015trpo,
  title={Trust Region Policy Optimization},
  author={Schulman, John and Levine, Sergey and Moritz, Philipp and Jordan, Michael I. and Abbeel, Pieter},
  journal={arXiv preprint arXiv:1502.05477},
  year={2015}
}

@article{schulman2017ppo,
  title={Proximal Policy Optimization Algorithms},
  author={Schulman, John and Wolski, Filip and Dhariwal, Prafulla and Radford, Alec and Klimov, Oleg},
  journal={arXiv preprint arXiv:1707.06347},
  year={2017}
}

@article{shao2024deepseekmath,
  title={DeepSeekMath: Pushing the Limits of Mathematical Reasoning in Open Language Models},
  author={Shao, Zhihong and Wang, Peiyi and Zhu, Qihao and Xu, Runxin and Song, Junxiao and Bi, Xiao and Zhang, Haowei and Zhang, Mingchuan and Li, Y. K. and Wu, Y. and Guo, Daya},
  journal={arXiv preprint arXiv:2402.03300},
  year={2024}
}

@article{sun2023llavarlhf,
  title={Aligning Large Multimodal Models with Factually Augmented RLHF},
  author={Sun, Zhiqing and Shen, Sheng and Cao, Shengcao and Liu, Haotian and Li, Chunyuan and Shen, Yikang and Gan, Chuang and Gui, Liang-Yan and Wang, Yu-Xiong and Yang, Yiming and Keutzer, Kurt and Darrell, Trevor},
  journal={arXiv preprint arXiv:2309.14525},
  year={2023}
}

@article{yu2023rlhfv,
  title={RLHF-V: Towards Trustworthy MLLMs via Behavior Alignment from Fine-grained Correctional Human Feedback},
  author={Yu, Tianyu and Yao, Yuan and Zhang, Haoye and He, Taiwen and Han, Yifeng and Cui, Ganqu and Hu, Jinyi and Liu, Zhiyuan and Zheng, Hai-Tao and Sun, Maosong and Chua, Tat-Seng},
  journal={arXiv preprint arXiv:2312.00849},
  year={2023}
}

@article{zhang2025mmrlhf,
  title={MM-RLHF: The Next Step Forward in Multimodal LLM Alignment},
  author={Zhang, Yi-Fan and Yu, Tao and Tian, Haochen and Fu, Chaoyou and Li, Peiyan and Zeng, Jianshu and Xie, Wulin and Shi, Yang and Zhang, Huanyu and Wu, Junkang and Wang, Xue and Hu, Yibo and Wen, Bin and Yang, Fan and Zhang, Zhang and Gao, Tingting and Zhang, Di and Wang, Liang and Jin, Rong and Tan, Tieniu},
  journal={arXiv preprint arXiv:2502.10391},
  year={2025}
}

@article{ji2025saferlhfv,
  title={Safe RLHF-V: Safe Reinforcement Learning from Multi-modal Human Feedback},
  author={Ji, Jiaming and Chen, Xinyu and Pan, Rui and Zhang, Conghui and Zhu, Han and Li, Jiahao and Hong, Donghai and Chen, Boyuan and Zhou, Jiayi and Wang, Kaile and Dai, Juntao and Chan, Chi-Min and Tang, Yida and Han, Sirui and Guo, Yike and Yang, Yaodong},
  journal={arXiv preprint arXiv:2503.17682},
  year={2025}
}

@article{liu2025visualrft,
  title={Visual-RFT: Visual Reinforcement Fine-Tuning},
  author={Liu, Ziyu and Sun, Zeyi and Zang, Yuhang and Dong, Xiaoyi and Cao, Yuhang and Duan, Haodong and Lin, Dahua and Wang, Jiaqi},
  journal={arXiv preprint arXiv:2503.01785},
  year={2025}
}

@article{shen2025vlmr1,
  title={VLM-R1: A Stable and Generalizable R1-style Large Vision-Language Model},
  author={Shen, Haozhan and Liu, Peng and Li, Jingcheng and Fang, Chunxin and Ma, Yibo and Liao, Jiajia and Shen, Qiaoli and Zhang, Zilun and Zhao, Kangjia and Zhang, Qianqian and Xu, Ruochen and Zhao, Tiancheng},
  journal={arXiv preprint arXiv:2504.07615},
  year={2025}
}

@article{zhang2025r1vl,
  title={R1-VL: Learning to Reason with Multimodal Large Language Models via Step-wise Group Relative Policy Optimization},
  author={Zhang, Jingyi and Huang, Jiaxing and Yao, Huanjin and Liu, Shunyu and Zhang, Xikun and Lu, Shijian and Tao, Dacheng},
  journal={arXiv preprint arXiv:2503.12937},
  year={2025}
}

@article{tan2025reasonrft,
  title={Reason-RFT: Reinforcement Fine-Tuning for Visual Reasoning},
  author={Tan, Huajie and Ji, Yuheng and Hao, Xiaoshuai and Lin, Minglan and Wang, Pengwei and Wang, Zhongyuan and Zhang, Shanghang},
  journal={arXiv preprint arXiv:2503.20752},
  year={2025}
}

@article{liu2025visualarft,
  title={Visual Agentic Reinforcement Fine-Tuning},
  author={Liu, Ziyu and Zang, Yuhang and Zou, Yushan and Liang, Zijian and Dong, Xiaoyi and Cao, Yuhang and Duan, Haodong and Lin, Dahua and Wang, Jiaqi},
  journal={arXiv preprint arXiv:2505.14246},
  year={2025}
}

@article{wang2025papo,
  title={Perception-Aware Policy Optimization for Multimodal Reasoning},
  author={Wang, Zhenhailong and Guo, Xuehang and Stoica, Sofia and Xu, Haiyang and Wang, Hongru and Ha, Hyeonjeong and Chen, Xiusi and Chen, Yangyi and Yan, Ming and Huang, Fei and Ji, Heng},
  journal={arXiv preprint arXiv:2507.06448},
  year={2025}
}

@article{chen2025pebr,
  title={Perception Before Reasoning: Two-Stage Reinforcement Learning for Visual Reasoning in Vision-Language Models},
  author={Chen, Yan and Li, Long and Xi, Teng and Zeng, Long and Wang, Jingdong},
  journal={arXiv preprint arXiv:2509.13031},
  year={2025}
}

@article{li2025visionsr1,
  title={Self-Rewarding Vision-Language Model via Reasoning Decomposition},
  author={Li, Zongxia and Yu, Wenhao and Huang, Chengsong and Liu, Rui and Liang, Zhenwen and Liu, Fuxiao and Chen, Jingxi and Yu, Dian and Boyd-Graber, Jordan and Mi, Haitao and Yu, Dong},
  journal={arXiv preprint arXiv:2508.19652},
  year={2025}
}

@article{sarch2025vigorl,
  title={Grounded Reinforcement Learning for Visual Reasoning},
  author={Sarch, Gabriel and Saha, Snigdha and Khandelwal, Naitik and Jain, Ayush and Tarr, Michael J. and Kumar, Aviral and Fragkiadaki, Katerina},
  journal={arXiv preprint arXiv:2505.23678},
  year={2025}
}

@article{hinton2015distilling,
  title={Distilling the Knowledge in a Neural Network},
  author={Hinton, Geoffrey and Vinyals, Oriol and Dean, Jeff},
  journal={arXiv preprint arXiv:1503.02531},
  year={2015}
}

@inproceedings{kim2016seqkd,
  title={Sequence-Level Knowledge Distillation},
  author={Kim, Yoon and Rush, Alexander M.},
  booktitle={Proceedings of the 2016 Conference on Empirical Methods in Natural Language Processing},
  year={2016}
}

@article{agarwal2024gkd,
  title={On-Policy Distillation of Language Models: Learning from Self-Generated Mistakes},
  author={Agarwal, Rishabh and Vieillard, Nino and Zhou, Yongchao and Stanczyk, Piotr and Ramos, Sabela and Geist, Matthieu and Bachem, Olivier},
  journal={arXiv preprint arXiv:2306.13649},
  year={2024}
}

@article{yang2026gopd,
  title={Learning beyond Teacher: Generalized On-Policy Distillation with Reward Extrapolation},
  author={Yang, Wenkai and Liu, Weijie and Xie, Ruobing and Yang, Kai and Yang, Saiyong and Lin, Yankai},
  journal={arXiv preprint arXiv:2602.12125},
  year={2026}
}

@article{ko2026reopold,
  title={Scaling Reasoning Efficiently via Relaxed On-Policy Distillation},
  author={Ko, Jongwoo and Abdali, Sara and Kim, Young Jin and Chen, Tianyi and Cameron, Pashmina},
  journal={arXiv preprint arXiv:2603.11137},
  year={2026}
}

@article{ye2026opcd,
  title={On-Policy Context Distillation for Language Models},
  author={Ye, Tianzhu and Dong, Li and Wu, Xun and Huang, Shaohan and Wei, Furu},
  journal={arXiv preprint arXiv:2602.12275},
  year={2026}
}

@article{zhang2026prefixopd,
  title={Fast and Effective On-policy Distillation from Reasoning Prefixes},
  author={Zhang, Dongxu and Yang, Zhichao and Janghorbani, Sepehr and Han, Jun and Ressler, Andrew and Qian, Qian and Lyng, Gregory D. and Batra, Sanjit Singh and Tillman, Robert E.},
  journal={arXiv preprint arXiv:2602.15260},
  year={2026}
}

@article{jin2026entropy,
  title={Entropy-Aware On-Policy Distillation of Language Models},
  author={Jin, Woogyeol and Min, Taywon and Yang, Yongjin and Kadhe, Swanand Ravindra and Zhou, Yi and Wei, Dennis and Baracaldo, Nathalie and Lee, Kimin},
  journal={arXiv preprint arXiv:2603.07079},
  year={2026}
}

@article{bai2025qwen25vl,
  title={Qwen2.5-VL Technical Report},
  author={Bai, Shuai and Chen, Keqin and Liu, Xuejing and Wang, Jialin and Ge, Wenbin and Song, Sibo and Dang, Kai and Wang, Peng and Wang, Shijie and Tang, Jun and others},
  journal={arXiv preprint arXiv:2502.13923},
  year={2025}
}

@article{wang2025vlrethinker,
  title={VL-Rethinker: Incentivizing Self-Reflection of Vision-Language Models with Reinforcement Learning},
  author={Wang, Haozhe and Qu, Chao and Huang, Zuming and Chu, Wei and Lin, Fangzhen and Chen, Wenhu},
  journal={arXiv preprint arXiv:2504.08837},
  year={2025}
}

@inproceedings{lu2021intergps,
  title={Inter-GPS: Interpretable Geometry Problem Solving with Formal Language and Symbolic Reasoning},
  author={Lu, Pan and Gong, Ran and Jiang, Shibiao and Qiu, Liang and Huang, Siyuan and Liang, Xiaodan and Zhu, Song-Chun},
  booktitle={Proceedings of the 59th Annual Meeting of the Association for Computational Linguistics},
  year={2021}
}

@inproceedings{lu2023mathvista,
  title={MathVista: Evaluating Mathematical Reasoning of Foundation Models in Visual Contexts},
  author={Lu, Pan and Bansal, Hritik and Xia, Tony and Liu, Jiacheng and Li, Chunyuan and Hajishirzi, Hannaneh and Cheng, Hao and Chang, Kai-Wei and Galley, Michel and Gao, Jianfeng},
  booktitle={International Conference on Learning Representations},
  year={2024}
}

@inproceedings{zhang2024mathverse,
  title={MathVerse: Does Your Multi-modal LLM Truly See the Diagrams in Visual Math Problems?},
  author={Zhang, Renrui and Jiang, Dongzhi and Zhang, Yichi and Lin, Haokun and Guo, Ziyu and Qiu, Pengshuo and Zhou, Aojun and Lu, Pan and Chang, Kai-Wei and Gao, Peng and Li, Hongsheng},
  booktitle={European Conference on Computer Vision},
  year={2024}
}

@article{qiao2024wemath,
  title={We-Math: Does Your Large Multimodal Model Achieve Human-like Mathematical Reasoning?},
  author={Qiao, Runqi and Tan, Qiuna and Dong, Guanting and Wu, Minhui and Sun, Chong and Song, Xiaoshuai and GongQue, Zhuoma and Lei, Shanglin and Wei, Zhe and Zhang, Miaoxuan and others},
  journal={arXiv preprint arXiv:2407.01284},
  year={2024}
}

@article{yue2024mmmupro,
  title={MMMU-Pro: A More Robust Multi-discipline Multimodal Understanding Benchmark},
  author={Yue, Xiang and Zheng, Tianyu and Ni, Yuansheng and Wang, Yubo and Zhang, Kai and Tong, Shengbang and Sun, Yuxuan and Yu, Botao and Zhang, Ge and Sun, Huan and Su, Yu and Chen, Wenhu and Neubig, Graham},
  journal={arXiv preprint arXiv:2409.02813},
  year={2024}
}

@article{xiao2024logicvista,
  title={LogicVista: Multimodal LLM Logical Reasoning Benchmark in Visual Contexts},
  author={Xiao, Yijia and Sun, Edward and Liu, Tianyu and Wang, Wei},
  journal={arXiv preprint arXiv:2407.04973},
  year={2024}
}

@inproceedings{li2023superclevr,
  title={Super-CLEVR: A Virtual Benchmark to Diagnose Domain Robustness in Visual Reasoning},
  author={Li, Zhenwen and Wang, Xuehui and Stengel-Eskin, Elias and Kortylewski, Adam and Ma, Wufei and Van Durme, Benjamin and Yuille, Alan},
  booktitle={Proceedings of the IEEE/CVF Conference on Computer Vision and Pattern Recognition},
  year={2023}
}

@article{Zheng2025DeepEyesI,
  title={DeepEyes: Incentivizing "Thinking with Images" via Reinforcement Learning},
  author={Ziwei Zheng and Michael Yang and Jack Hong and Chen-Xiao Zhao and Guo-Hai Xu and Le Yang and Chao Shen and Xing Yu},
  journal={ArXiv},
  year={2025},
  volume={abs/2505.14362},
  url={https://api.semanticscholar.org/CorpusID:278769859}
}

@article{Zhang2025ThymeTB,
  title={Thyme: Think Beyond Images},
  author={Yi-Fan Zhang and Xingyu Lu and Shu-Kang Yin and Chaoyou Fu and Wei Chen and Xiao Hu and Bin Wen and Kai-Yu Jiang and Chang-Yi Liu and Tian-Ke Zhang and Hao-Nan Fan and Kaibing Chen and Jian-Kang Chen and Haojie Ding and Kaiyu Tang and Zhang Zhang and Liang Wang and Fan Yang and Ting-Ting Gao and Guorui Zhou},
  journal={ArXiv},
  year={2025},
  volume={abs/2508.11630},
  url={https://api.semanticscholar.org/CorpusID:280671726}
}

@article{Bai2025Qwen3VLTR,
  title={Qwen3-VL Technical Report},
  author={Shuai Bai and Yuxuan Cai and Rui-Zhe Chen and Ke-qin Chen and Xiong-Hui Chen and Zesen Cheng and Liang-Hao Deng and Wei Ding and Rongyao Fang and Chang Gao and Chunjiang Ge and others},
  journal={ArXiv},
  year={2025},
  volume={abs/2511.21631},
  url={https://api.semanticscholar.org/CorpusID:283262018}
}

@article{Wang2025InternVL35AO,
  title={InternVL3.5: Advancing Open-Source Multimodal Models in Versatility, Reasoning, and Efficiency},
  author={Wei-Yun Wang and Zhangwei Gao and Lixin Gu and Hengjun Pu and Long Cui and Xingguang Wei and Zhaoyang Liu and Linglin Jing and Sheng-Long Ye and Jie Shao and Zhaokai Wang and Zhe Chen and Hongjie Zhang and others},
  journal={ArXiv},
  year={2025},
  volume={abs/2508.18265},
  url={https://api.semanticscholar.org/CorpusID:280710824}
}
\normalsize

\newpage
\appendix

\section{Additional theoretical analysis}
\label{app:theory}

This appendix provides additional theoretical justification for VC-OPD. 
The main text already shows that OPD can be interpreted as token-level policy optimization induced by the reverse KL divergence. 
Here we focus on three properties that are specific to VC-OPD: 
(i) the factual--counterfactual teacher gap is a pointwise visual-dependence signal, 
(ii) mass-preserving normalization stabilizes the total update scale while preserving relative token importance, and 
(iii) the counterfactual branch modulates update magnitude without changing the update direction.

\subsection{Counterfactual teacher gap as visual likelihood ratio}

Let $s_t=(x,v,y_{<t})$ and $a_t=y_t$. 
For the same sampled token, VC-OPD evaluates the teacher under the factual visual input and the counterfactual visual input:
\begin{equation}
p_T^{+}(a_t)=p_T(a_t\mid x,v,y_{<t}), 
\qquad
p_T^{-}(a_t)=p_T(a_t\mid x,\tilde v,y_{<t}).
\end{equation}
The visual dependence score used in VC-OPD is
\begin{equation}
\Delta_t^{\mathrm{vis}}
=
\log p_T^{+}(a_t)-\log p_T^{-}(a_t)
=
\log
\frac{p_T(a_t\mid x,v,y_{<t})}
     {p_T(a_t\mid x,\tilde v,y_{<t})}.
\label{eq:app_visual_ratio_compact}
\end{equation}
Thus, $\Delta_t^{\mathrm{vis}}$ is a pointwise log-likelihood ratio between the factual and counterfactual teacher distributions. 
A positive value means that the real image increases the teacher's support for the token relative to the counterfactual image, while a value close to zero means that the token is largely invariant to the removal of instance-specific visual evidence. 
This justifies using $\Delta_t^{\mathrm{vis}}$ as a token-level grounding signal.

VC-OPD converts this score into a non-negative soft weight:
\begin{equation}
w_t=
\sigma\!\left(
\frac{\Delta_t^{\mathrm{vis}}-\tau_v}{T_v}
\right).
\label{eq:app_soft_weight_compact}
\end{equation}
Since the sigmoid function is monotonic, if 
$\Delta_i^{\mathrm{vis}}>\Delta_j^{\mathrm{vis}}$, then $w_i>w_j$. 
Therefore, the soft weighting function preserves the ranking induced by the visual likelihood ratio: tokens more dependent on real visual evidence receive larger update weights.

\subsection{Mass-preserving normalization}

Directly using $w_t$ may make the total update scale vary across responses, because different examples contain different numbers of visually grounded tokens. 
VC-OPD therefore uses the normalized weight
\begin{equation}
\bar{w}_t
=
m_t w_t
\cdot
\frac{N}{\sum_{j=1}^{T}m_jw_j+\epsilon},
\qquad
N=\sum_{t=1}^{T}m_t.
\end{equation}
where $m_t$ is the response mask. 
Ignoring the small numerical constant $\epsilon$, the normalized weights satisfy
\begin{equation}
\sum_{t=1}^{T}\bar w_t = N.
\label{eq:app_mass_conservation_compact}
\end{equation}
Thus, the total token weight is approximately equal to the number of valid response tokens. 
At the same time, for any two valid tokens $i$ and $j$,
\begin{equation}
\frac{\bar w_i}{\bar w_j}
=
\frac{w_i}{w_j}.
\label{eq:app_relative_preservation_compact}
\end{equation}
Therefore, mass-preserving normalization keeps the relative visual emphasis unchanged while stabilizing the overall update magnitude. 
This property explains why VC-OPD can emphasize visually grounded tokens without changing the effective learning scale across batches.

\subsection{Direction--magnitude separation}

The final VC-OPD advantage is
\begin{equation}
A_t^{\mathrm{VC}}
=
\bar w_t \cdot \tilde A_t^{\mathrm{dir}},
\qquad
A_t^{\mathrm{dir}}
=
\log p_T^{+}(a_t)-\log \mu(a_t\mid s_t),
\label{eq:app_final_adv_compact}
\end{equation}
where $\tilde A_t^{\mathrm{dir}}$ is a normalized version of the factual teacher--student direction term. 
The resulting objective is
\begin{equation}
\mathcal{L}_{\mathrm{VC\text{-}OPD}}(\theta)
=
-
\mathbb{E}_{y\sim\mu}
\left[
\sum_{t=1}^{T}
\mathrm{sg}\!\left(A_t^{\mathrm{VC}}\right)
\log \pi_\theta(a_t\mid s_t)
\right].
\label{eq:app_vcopd_loss_compact}
\end{equation}
Taking the gradient gives
\begin{equation}
\nabla_\theta \mathcal{L}_{\mathrm{VC\text{-}OPD}}
=
-
\mathbb{E}_{y\sim\mu}
\left[
\sum_{t=1}^{T}
\mathrm{sg}\!\left(\bar w_t\tilde A_t^{\mathrm{dir}}\right)
\nabla_\theta \log \pi_\theta(a_t\mid s_t)
\right].
\label{eq:app_vcopd_grad_compact}
\end{equation}
This is still an advantage-weighted policy-gradient estimator, but the advantage is reweighted by token-level visual dependence.

Since $\bar w_t\geq 0$, the counterfactual branch cannot reverse the update direction:
\begin{equation}
\mathrm{sign}\!\left(\bar w_t A_t^{\mathrm{dir}}\right)
=
\mathrm{sign}\!\left(A_t^{\mathrm{dir}}\right)
\quad
\text{whenever } \bar w_t>0.
\label{eq:app_sign_preserve_compact}
\end{equation}
Therefore, the factual teacher--student discrepancy determines whether the student should increase or decrease the probability of a sampled token, while the counterfactual teacher only determines how strongly this update should be applied. 
This separation is important for stability: the corrupted visual input is used to estimate grounding strength, but it does not directly decide the sign of the optimization signal.

\subsection{View as selective token-level distillation}

VC-OPD can also be interpreted as performing OPD under a reweighted token distribution. 
For a fixed response, define
\begin{equation}
q(t\mid y)
=
\frac{m_t w_t}{\sum_{j=1}^{T}m_jw_j}.
\label{eq:app_token_distribution_compact}
\end{equation}
Then the VC-OPD objective is proportional to
\begin{equation}
-
\mathbb{E}_{y\sim\mu}
\left[
N\cdot
\mathbb{E}_{t\sim q(\cdot\mid y)}
\left[
\mathrm{sg}\!\left(\tilde A_t^{\mathrm{dir}}\right)
\log \pi_\theta(a_t\mid s_t)
\right]
\right].
\label{eq:app_selective_distill_compact}
\end{equation}
Thus, VC-OPD can be viewed as standard token-level OPD applied to a token distribution biased toward visually grounded tokens, while preserving the total update mass. 
This interpretation formalizes the main design goal of VC-OPD: retain the dense supervision advantage of OPD, but reallocate learning capacity toward tokens that are supported by real visual evidence.

\section{Training Details}
\label{app:training_details}

\subsection{Stage-I warm-start}

Before visual counterfactual on-policy distillation, we first warm-start the student with teacher-generated responses. 
For each training example, the teacher is prompted to produce a reasoning response followed by a boxed final answer. 
The prompt template is:

\begin{quote}
\small
\ttfamily
\{\{ content \textbar{} trim \}\}\\
You FIRST think about the reasoning process as an internal monologue and then provide the final answer. 
The reasoning process MUST BE enclosed within <think> </think> tags. 
The final answer MUST BE put in \textbackslash boxed\{\}.
\end{quote}

After generation, we extract the final boxed answer and compare it with the ground-truth answer using the same answer-extraction and exact-match evaluation protocol as in the main experiments. 
Only examples for which the teacher produces a correct final answer are retained for warm-start training. 
This filtering removes noisy teacher trajectories and ensures that the student is initialized with reliable response formats and valid reasoning paths before entering on-policy optimization.

The warm-start stage uses standard autoregressive supervision on the filtered teacher responses. 
Table~\ref{tab:warm_start_hparams} summarizes the warm-start hyperparameters.

\begin{table}[t]
\centering
\small
\setlength{\tabcolsep}{8pt}
\renewcommand{\arraystretch}{1.12}
\caption{Warm-start training hyperparameters.}
\label{tab:warm_start_hparams}
\begin{tabular}{lc}
\toprule
\textbf{Hyperparameter} & \textbf{Value} \\
\midrule
Student model & Qwen2.5-VL-3B-Instruct \\
Teacher model & Qwen2.5-VL-32B-Instruct  \\
Optimizer & AdamW \\
Learning rate & $1 \times 10^{-6}$ \\
Batch size & 256 \\
Response format & \texttt{<think>} reasoning + \texttt{\textbackslash boxed\{\}} answer \\
\bottomrule
\end{tabular}
\end{table}

\subsection{Stage-II on-policy training}

After warm-start, we perform on-policy optimization using the same training prompts. 
GRPO is trained with verifiable outcome-level rewards, while VC-OPD uses the student-generated trajectories for teacher scoring under both factual and counterfactual visual inputs. 
The stage-II implementation is based on the \texttt{verl} training framework. 
For VC-OPD, the factual teacher branch provides the token-level distillation direction, while the counterfactual teacher branch estimates visual dependence and reweights the token-level objective.

Table~\ref{tab:stage2_hparams} reports the main hyperparameters used for GRPO and VC-OPD.

\begin{table}[t]
\centering
\small
\setlength{\tabcolsep}{7pt}
\renewcommand{\arraystretch}{1.12}
\caption{Stage-II training hyperparameters for GRPO and VC-OPD.}
\label{tab:stage2_hparams}
\begin{tabular}{lcc}
\toprule
\textbf{Hyperparameter} & \textbf{GRPO} & \textbf{VC-OPD} \\
\midrule
Optimizer & AdamW & AdamW \\
Learning rate & $1 \times 10^{-6}$ & $5 \times 10^{-6}$ \\
Batch size (rollouts) & 5 & 1 \\
Max response length & 8192 & 8192 \\
Max prompt length & 4096 & 4096 \\
Generation temperature & 1.0 & 1.0 \\
Top-$p$ & 1.0 & 1.0 \\
Tensor parallel size & 2 & 2 \\
Visual token ratio & -- & 10\% \\
Counterfactual corruption & -- & Patch dropout 60\% \\

\bottomrule
\end{tabular}
\end{table}

\section{Log-probability Diagnostics}
\label{app:logprob_diagnostics}

We further visualize the log-probability traces during VC-OPD training to verify that the factual and counterfactual teacher scoring remains numerically stable. 
At each optimization step, we record the average response-token log probability for the student model, the factual teacher, and the counterfactual teacher. 
The light curve denotes the raw step-wise value, and the darker curve denotes the smoothed trend.

\begin{figure*}[t]
    \centering
    \begin{subfigure}[t]{0.32\textwidth}
        \centering
        \includegraphics[width=\linewidth]{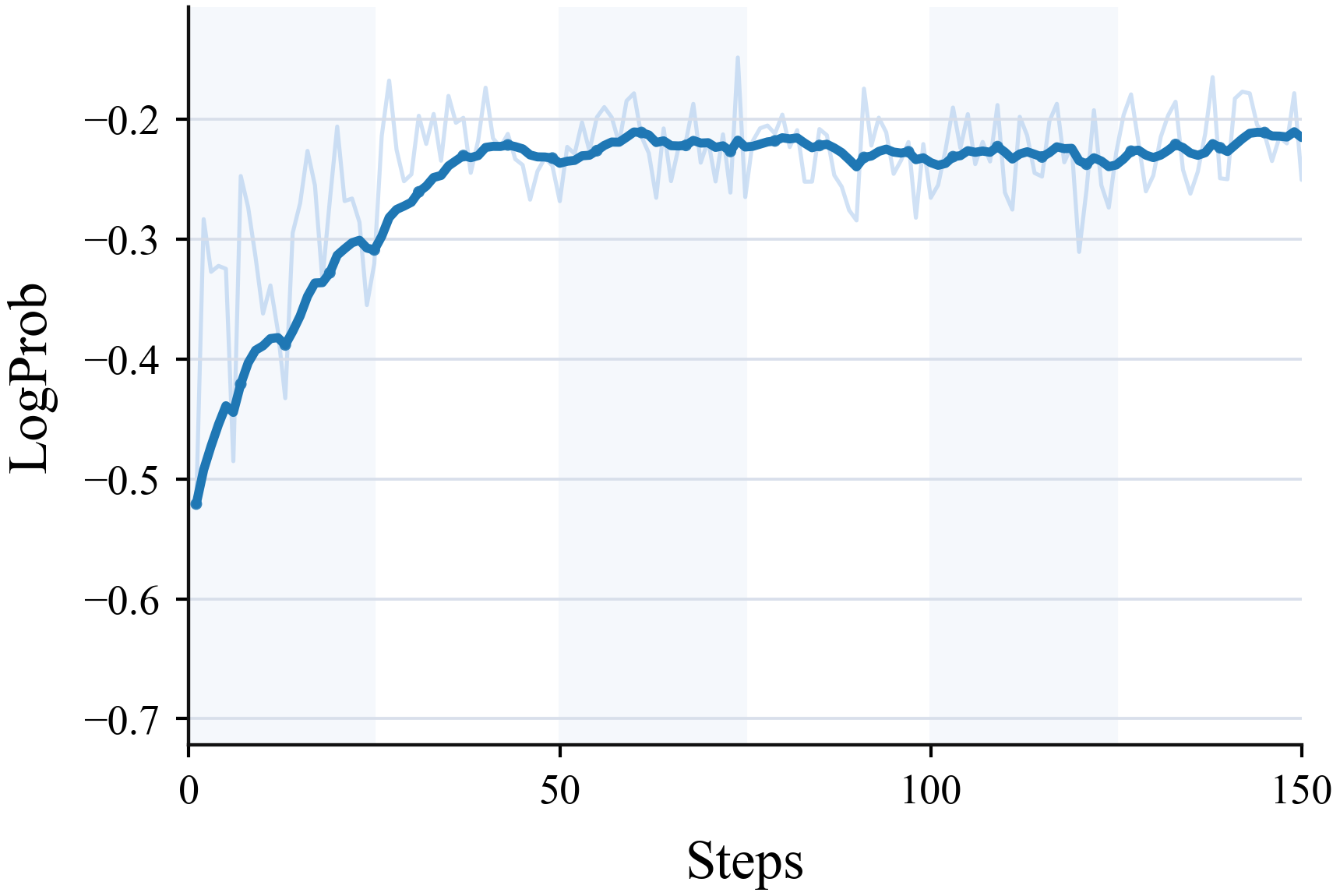}
        \caption{Student model}
        \label{fig:student_logprob}
    \end{subfigure}
    \hfill
    \begin{subfigure}[t]{0.32\textwidth}
        \centering
        \includegraphics[width=\linewidth]{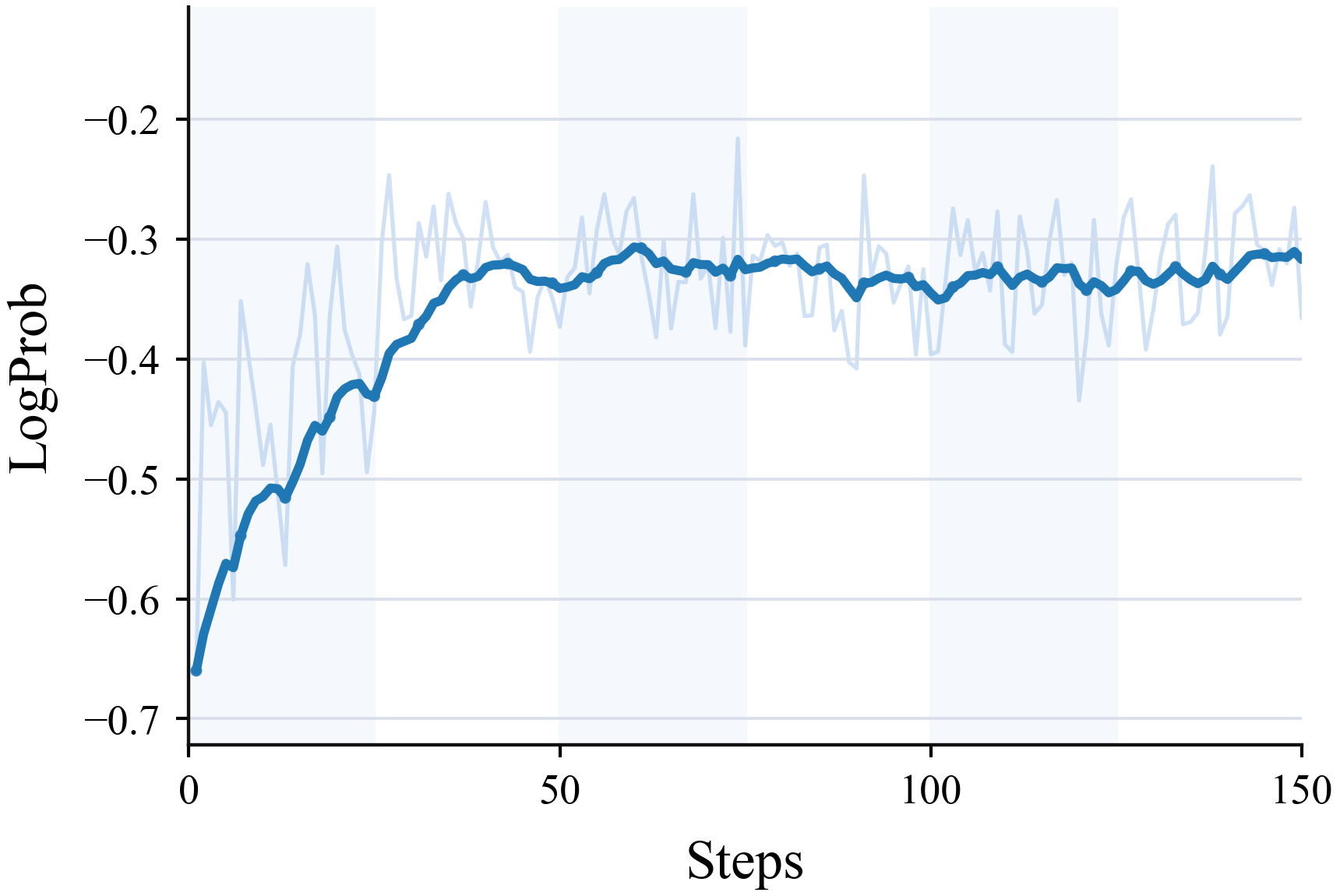}
        \caption{Factual teacher}
        \label{fig:factual_teacher_logprob}
    \end{subfigure}
    \hfill
    \begin{subfigure}[t]{0.32\textwidth}
        \centering
        \includegraphics[width=\linewidth]{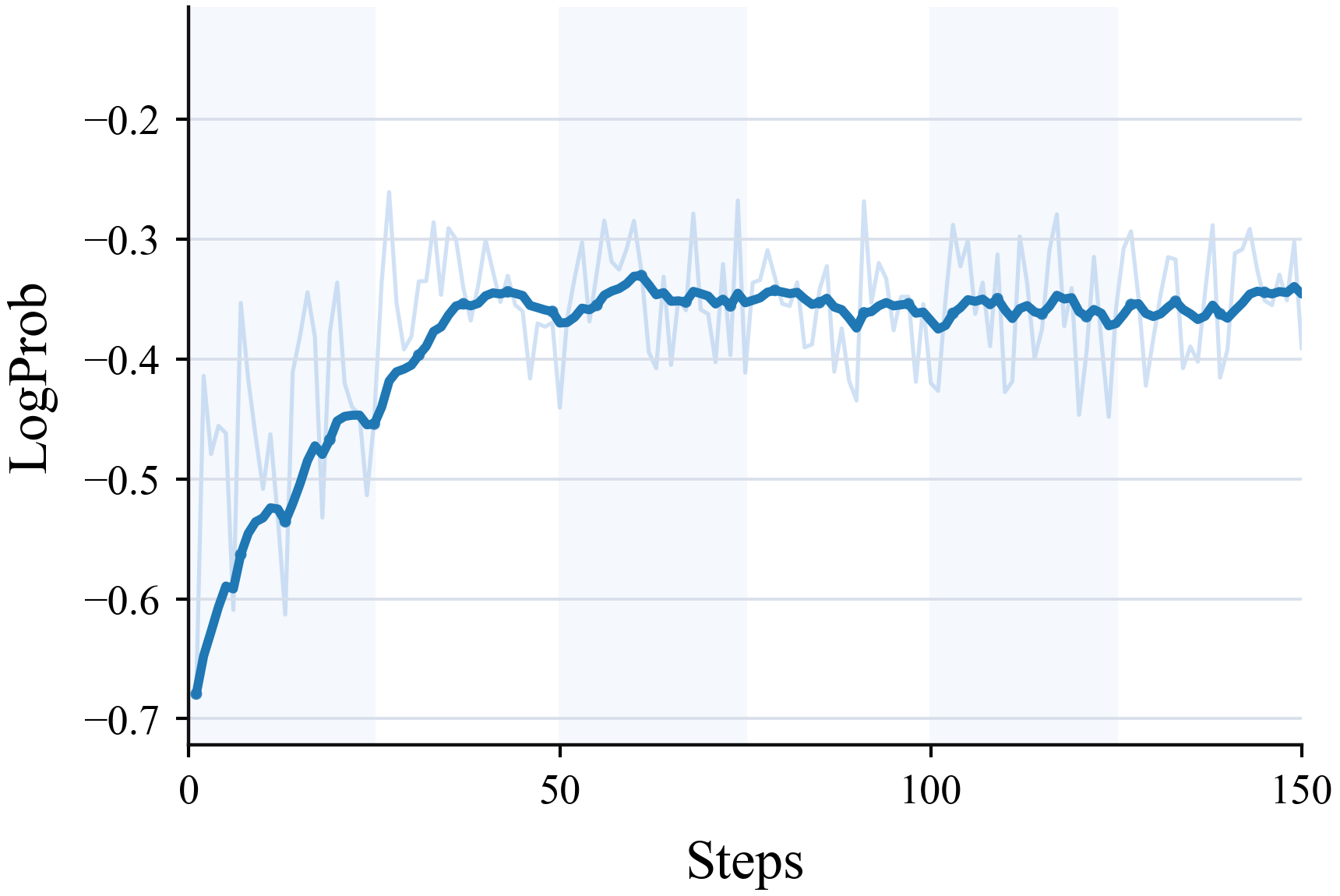}
        \caption{Counterfactual teacher}
        \label{fig:counterfactual_teacher_logprob}
    \end{subfigure}
    \caption{
    Log-probability diagnostics during VC-OPD training.
    We plot the average response-token log probability of the student, factual teacher, and counterfactual teacher over training steps.
    The curves stabilize after the early optimization phase, indicating that both teacher-scoring branches remain in a comparable and numerically stable prediction regime.
    }
    \label{fig:logprob_diagnostics}
\end{figure*}

As shown in Figure~\ref{fig:logprob_diagnostics}, the student log probability increases rapidly in the early stage and then gradually stabilizes, reflecting the effect of warm-starting followed by on-policy adaptation. 
The factual and counterfactual teacher branches remain stable throughout training, which is important because VC-OPD relies on their log-probability difference to estimate token-level visual dependence. 
The counterfactual branch does not serve as an independent training target; instead, it provides a relative grounding signal by measuring how teacher support changes when instance-specific visual evidence is weakened. 
These diagnostics suggest that the counterfactual scoring procedure does not introduce unstable probability estimates and can be used reliably for grounding-aware token reweighting.

\section{Token-level Grounding Verification}

The qualitative visualization in Figure~\ref{fig:token_grounding_visualization} suggests that the
factual--counterfactual teacher gap more selectively highlights image-dependent
tokens than the standard student--teacher gap. To quantitatively verify this
mechanism, we conduct a token-level grounding verification study on a subset of
validation examples.

We sample model responses from the validation set and annotate response tokens as
visually grounded if their correctness depends on image-specific evidence, such as
objects, quantities, spatial relations, diagram symbols, chart elements, or
answer-relevant visual attributes. Tokens that only serve as discourse markers,
generic reasoning templates, or language-prior continuations are annotated as
non-visual. Given these annotations, we compare four token scoring signals:
\begin{itemize}
    \item \textbf{Random score}: a random token ranking baseline.
    \item \textbf{Token log-probability}: the factual teacher log-probability of each token.
    \item \textbf{Student--teacher gap}: the standard OPD score based on the teacher--student log-probability difference.
    \item \textbf{Factual--counterfactual gap}: the proposed visual-dependence score used by VC-OPD.
\end{itemize}

For each scoring signal, we rank all response tokens within each example and
evaluate how well the top-ranked tokens match the human visual-grounding
annotations. We report Precision@5\%, Precision@10\%, Recall@10\%, and AUPRC.
These metrics measure whether a scoring signal can retrieve visually grounded
tokens from the full response.

As shown in Table~\ref{tab:token_grounding}, the factual--counterfactual gap
provides the strongest alignment with human-annotated visually grounded tokens.
The random baseline roughly reflects the sparsity of visually grounded tokens,
while token log-probability performs only slightly better because high-probability
tokens often correspond to common function words or response templates. The
student--teacher gap improves over these baselines, but it still ranks many
generic reasoning tokens highly because it measures teacher--student disagreement
rather than visual dependence. In contrast, the factual--counterfactual gap
achieves the highest precision and recall, indicating that it more accurately
prioritizes tokens whose correctness depends on image-specific evidence. This
result supports the central mechanism of VC-OPD: the counterfactual teacher branch
does not merely identify high-discrepancy tokens, but provides a token-level signal
that is more closely aligned with visual grounding.

\begin{table}[t]
    \centering
    \small
    \setlength{\tabcolsep}{6pt}
    \renewcommand{\arraystretch}{1.12}
    \caption{
        Token-level grounding verification.
        We compare different token scoring signals for identifying human-annotated
        visually grounded tokens.
    }
    \label{tab:token_grounding}
    \begin{tabular}{lcccc}
        \toprule
        \textbf{Scoring signal} &
        \textbf{P@5\%} &
        \textbf{P@10\%} &
        \textbf{Recall@10\%} &
        \textbf{AUPRC} \\
        \midrule
        Random score
        & 11.8 & 11.6 & 10.1 & 12.0 \\
        Token log-probability
        & 18.7 & 17.4 & 15.2 & 18.6 \\
        Student--teacher gap
        & 36.5 & 31.8 & 27.7 & 34.2 \\
        Factual--counterfactual gap
        & \textbf{58.9} & \textbf{52.1} & \textbf{45.4} & \textbf{57.6} \\
        \bottomrule
    \end{tabular}
\end{table}

\section{Generalization to Qwen3-VL}
\label{app:Qwen3-VL}
To evaluate whether VC-OPD transfers to newer VLM backbones, we further conduct experiments
with Qwen3-VL-2B~\cite{Bai2025Qwen3VLTR} and Qwen3-VL-4B students using Qwen3-VL-8B as the teacher.
RKL and VC-OPD use the same student initialization, training data, rollout budget,
and evaluation protocol. We additionally report the performance of the original base
models to provide a direct reference for the effectiveness of post-training.

\begin{table*}[t]
    \centering
    \small
    \setlength{\tabcolsep}{4.2pt}
    \renewcommand{\arraystretch}{1.12}
    \caption{
        Generalization results on the Qwen3-VL family.
        We report exact-match accuracy (\%) under the same evaluation protocol as the main experiments.
    }
    \label{tab:qwen3}
    \resizebox{\textwidth}{!}{
    \begin{tabular}{lcccccccc}
        \toprule
        \textbf{Model} & \textbf{Geo3K} & \textbf{MVer.} & \textbf{MVist.}
        & \textbf{WeM.} & \textbf{MMMU} & \textbf{Logic} & \textbf{Count} & \textbf{Avg.} \\
        \midrule

        \multicolumn{9}{l}{\textit{Qwen3-VL-2B}} \\
        \quad + Base
        & 28.20 & 39.50 & 59.80 & 56.60 & 27.50 & 36.80 & 55.10 & 43.36 \\
        \quad + RKL
        & 39.50 & 42.80 & 63.00 & 60.50 & 31.40 & 40.90 & 61.20 & 48.47 \\
        \quad + VC-OPD
        & \textbf{42.40} & \textbf{44.20} & \textbf{64.50} & \textbf{61.90}
        & \textbf{33.80} & \textbf{42.80} & \textbf{63.80} & \textbf{50.49} \\
        \quad $\Delta_{\mathrm{RKL}}$ (\%)
        & \pos{+7.34} & \pos{+3.27} & \pos{+2.38} & \pos{+2.31}
        & \pos{+7.64} & \pos{+4.65} & \pos{+4.25} & \pos{+4.16} \\

        \midrule

        \multicolumn{9}{l}{\textit{Qwen3-VL-4B}} \\
        \quad + Base
        & 35.20 & 42.80 & 64.00 & 61.80 & 32.20 & 41.50 & 62.00 & 48.50 \\
        \quad + RKL
        & 43.20 & 45.50 & 66.70 & 64.50 & 35.60 & 44.80 & 67.30 & 52.51 \\
        \quad + VC-OPD
        & \textbf{46.70} & \textbf{47.80} & \textbf{68.30} & \textbf{66.40}
        & \textbf{38.40} & \textbf{47.50} & \textbf{70.40} & \textbf{55.07} \\
        \quad $\Delta_{\mathrm{RKL}}$ (\%)
        & \pos{+8.10} & \pos{+5.05} & \pos{+2.40} & \pos{+2.95}
        & \pos{+7.87} & \pos{+6.03} & \pos{+4.61} & \pos{+4.87} \\

        \bottomrule
    \end{tabular}}
\end{table*}

VC-OPD consistently improves over RKL for both Qwen3-VL student scales.
For the 2B and 4B students, VC-OPD achieves average accuracies of $50.49$ and
$55.07$, corresponding to relative improvements of $4.16\%$ and $4.87\%$ over RKL,
respectively. Compared with the corresponding base models, the gains are also
substantial, with absolute improvements of $7.13$ and $6.57$ points in average
accuracy. These results indicate that the proposed visual-dependent token
reweighting remains effective when transferred to the newer Qwen3-VL family.

\section{Generalization to InternVL3.5}
\label{app:InternVL3.5}
To assess cross-family generalization, we further implement VC-OPD on the
VERL–InternVL framework and evaluate InternVL3.5-2B~\cite{Wang2025InternVL35AO} and InternVL3.5-4B students
with an InternVL3.5-8B teacher. RKL and VC-OPD use the same student initialization,
training data, rollout budget, and evaluation protocol. We also include the
corresponding base-model performance as a reference for evaluating the effect
of post-training.

\begin{table*}[t]
    \centering
    \small
    \setlength{\tabcolsep}{4.2pt}
    \renewcommand{\arraystretch}{1.12}
    \caption{
        Generalization results on the InternVL3.5 family.
        We report exact-match accuracy (\%) under the same evaluation protocol as the main experiments.
    }
    \label{tab:internvl}
    \resizebox{\textwidth}{!}{
    \begin{tabular}{lcccccccc}
        \toprule
        \textbf{Model} & \textbf{Geo3K} & \textbf{MVer.} & \textbf{MVist.}
        & \textbf{WeM.} & \textbf{MMMU} & \textbf{Logic} & \textbf{Count} & \textbf{Avg.} \\
        \midrule

        \multicolumn{9}{l}{\textit{InternVL3.5-2B}} \\
        \quad + Base
        & 30.80 & 39.40 & 60.50 & 28.40 & 31.20 & 41.00 & 56.20 & 41.07 \\
        \quad + RKL
        & 37.90 & 42.10 & 64.90 & 35.20 & 34.00 & 39.80 & 61.00 & 44.99 \\
        \quad + VC-OPD
        & \textbf{40.80} & \textbf{44.20} & \textbf{66.10} & \textbf{37.40}
        & \textbf{35.80} & \textbf{42.30} & \textbf{63.10} & \textbf{47.10} \\
        \quad $\Delta_{\mathrm{RKL}}$ (\%)
        & \pos{+7.65} & \pos{+4.99} & \pos{+1.85} & \pos{+6.25}
        & \pos{+5.29} & \pos{+6.28} & \pos{+3.44} & \pos{+4.70} \\

        \midrule

        \multicolumn{9}{l}{\textit{InternVL3.5-4B}} \\
        \quad + Base
        & 39.20 & 49.80 & 71.10 & 35.80 & 38.80 & 53.20 & 64.80 & 50.39 \\
        \quad + RKL
        & 44.70 & 53.10 & 73.50 & 47.00 & 42.00 & 54.10 & 68.50 & 54.70 \\
        \quad + VC-OPD
        & \textbf{47.30} & \textbf{55.00} & \textbf{74.40} & \textbf{49.10}
        & \textbf{44.30} & \textbf{56.00} & \textbf{71.20} & \textbf{56.76} \\
        \quad $\Delta_{\mathrm{RKL}}$ (\%)
        & \pos{+5.82} & \pos{+3.58} & \pos{+1.22} & \pos{+4.47}
        & \pos{+5.48} & \pos{+3.51} & \pos{+3.94} & \pos{+3.76} \\

        \bottomrule
    \end{tabular}}
\end{table*}

VC-OPD consistently outperforms RKL for both InternVL3.5 student scales,
achieving relative improvements of $4.70\%$ and $3.76\%$ in average accuracy
for the 2B and 4B students, respectively. Relative to the corresponding base
models, VC-OPD further improves average accuracy by $6.03$ and $6.37$ points,
respectively. The consistent gains across both student scales and a different
VLM family demonstrate that VC-OPD is not restricted to the Qwen architecture
used in the main experiments, and supports its cross-family generalization.

\section{Training Efficiency}

VC-OPD evaluates the teacher under both factual and counterfactual visual conditions.
The two conditions share the same student-generated trajectory and are concatenated
along the batch dimension for a single batched teacher invocation.
In our implementation, the additional counterfactual branch increases the wall-clock
time per optimization step by approximately $20\%$, rather than doubling the
end-to-end training cost.

Table~\ref{tab:time_accuracy} compares the wall-clock time required by RKL and VC-OPD
to reach the same target Geo3K accuracy.
Although VC-OPD introduces additional teacher computation at each optimization step,
it consistently reaches the reported accuracy targets earlier than RKL.
The difference becomes particularly evident at higher accuracy levels: VC-OPD reaches
$32\%$ accuracy in $7.2$ hours compared with $14.5$ hours for RKL, and further reaches
$33\%$ and $34\%$ in $14.4$ and $17.4$ hours, respectively, whereas RKL does not
reach these targets within the reported training budget.
These results indicate that the extra per-step computation can be offset by faster
optimization progress, leading to improved time-to-accuracy.

\begin{table}[t]
    \centering
    \small
    \renewcommand{\arraystretch}{1.15}
    \caption{
        Wall-clock time (hours) required to reach target Geo3K accuracies for Qwen2.5-VL-3B.
    }
    \label{tab:time_accuracy}
    \begin{tabular*}{0.92\textwidth}{@{\extracolsep{\fill}}lcccccc@{}}
        \toprule
        \textbf{Target Accuracy} & \textbf{29\%} & \textbf{30\%} & \textbf{31\%} &
        \textbf{32\%} & \textbf{33\%} & \textbf{34\%} \\
        \midrule
        RKL    & 2.5 & 4.0 & 7.0 & 14.5 & -- & -- \\
        VC-OPD & \textbf{1.5} & \textbf{2.8} & \textbf{5.0} & \textbf{7.2}  & \textbf{14.4} & \textbf{17.4} \\
        \bottomrule
    \end{tabular*}
\end{table}

\section{Comparison with Recent Visual Reasoning Methods}

We additionally compare the Qwen2.5-VL-7B student distilled from the
Qwen2.5-VL-32B teacher with recent visual reasoning post-training methods on
four shared benchmarks. Since DeepEyes~\cite{Zheng2025DeepEyesI} and
Thyme-VL~\cite{Zhang2025ThymeTB} report results obtained under potentially
different training data, evaluation splits, and tool-augmented inference
procedures, this comparison should be interpreted as an external reference
rather than a strictly controlled comparison.


Table~\ref{tab:recent_methods} reports the results on MathVista, MathVerse, We-Math,
and LogicVista.
VC-OPD achieves an average accuracy of $59.13\%$ across the four benchmarks.
Compared with the reported results of DeepEyes and Thyme-VL, VC-OPD shows particularly
large gains on We-Math, while the differences on the other benchmarks are more modest.
Because the compared methods may differ in training resources, inference settings,
and evaluation protocols, these results are provided primarily to contextualize
the empirical performance of VC-OPD rather than to establish a controlled
state-of-the-art comparison.

\begin{table}[t]
\centering
\small
\renewcommand{\arraystretch}{1.12}
\caption{
Comparison with recent visual reasoning methods on shared benchmarks.
}
\label{tab:recent_methods}
\begin{tabular*}{0.92\textwidth}{@{\extracolsep{\fill}}lccccc@{}}
\toprule
\textbf{Method}
& \textbf{MathVista}
& \textbf{MathVerse}
& \textbf{We-Math}
& \textbf{LogicVista}
& \textbf{Avg.} \\
\midrule
DeepEyes
& 70.10 & 47.30 & 38.90 & 47.70 & 51.00 \\
Thyme-VL
& 70.00 & 39.10 & 39.30 & \textbf{49.00} & 49.35 \\
VC-OPD
& \textbf{70.90} & \textbf{48.90} & \textbf{68.10} & 48.60 & \textbf{59.13} \\
\bottomrule
\end{tabular*}
\end{table}


\end{document}